\documentclass[final]{l4dc2025}

\usepackage{algorithmic}
\usepackage{algorithm}
\usepackage{times}
\usepackage[utf8]{inputenc} 
\usepackage[T1]{fontenc}    
\usepackage{hyperref}       
\usepackage{url}            
\usepackage{booktabs}       
\usepackage{amsfonts}       
\usepackage{amsmath}
\usepackage{amssymb}
\usepackage{bm}
\usepackage{nicefrac}       
\usepackage{microtype}      
\usepackage{xcolor}         
\usepackage{graphicx}
\usepackage[font=small]{caption}
\usepackage{array}
\usepackage{cleveref}
\usepackage{rotating}
\usepackage{wrapfig}
\usepackage{enumitem}
\usepackage{times}
\usepackage{comment}

\begin{document}

\title[Continual Lifting of Koopman Dynamics]{Continual Learning and Lifting of Koopman Dynamics\\ for Linear Control of Legged Robots}

\coltauthor{%
 \Name{Feihan Li} \Email{feihanl@andrew.cmu.edu}\\
 \Name{Abulikemu Abuduweili} \Email{abulikea@andrew.cmu.edu}\\
 \Name{Yifan Sun} \Email{yifansu2@andrew.cmu.edu}\\
 \Name{Rui Chen} \Email{ruic3@andrew.cmu.edu} \\
 \Name{Weiye Zhao} \Email{weiyezha@andrew.cmu.edu} \\
 \Name{Changliu Liu} \Email{cliu6@andrew.cmu.edu} \\
 \addr Robotics Institute, Carnegie Mellon University, Pittsburgh, PA 15213, USA
}



\maketitle

\begin{abstract}
The control of legged robots, particularly humanoid and quadruped robots, presents significant challenges due to their high-dimensional and nonlinear dynamics. While linear systems can be effectively controlled using methods like Model Predictive Control (MPC), the control of nonlinear systems remains complex. One promising solution is the Koopman Operator, which approximates nonlinear dynamics with a linear model, enabling the use of proven linear control techniques. However, achieving accurate linearization through data-driven methods is difficult due to approximation error and domain shifts. These challenges restrict the scalability of Koopman-based approaches. This paper addresses these challenges by proposing an Incremental Koopman algorithm designed to iteratively refine Koopman dynamics for high-dimensional legged robots. The key idea is to progressively expand the dataset and latent space dimension, enabling the learned Koopman dynamics to converge towards true system dynamics. Theoretical analysis shows that the linear approximation error of our method converges monotonically. Experimental results demonstrate its superiority on robots like  Unitree G1, H1, A1, Go2, and ANYmal D, across various terrains. 
This is the first work to apply linearized whole body dynamics with the Koopman Operator for locomotion control of high-dimensional legged robots, providing a scalable model-based control solution.
The code can be found at: \url{https://github.com/intelligent-control-lab/Incremental-Koopman}.
\end{abstract}


\section{Introduction}

Control of legged robots has attracted growing interest, driven by the potential of humanoid and quadruped robots in real-world applications. However, the task remains challenging due to the high-dimensional, nonlinear dynamics of legged locomotion. Control strategies for nonlinear systems face a trade-off between controller complexity and modeling effort, broadly categorized as model-free or model-based.
Model-free methods directly derive the control law without explicitly modeling the system dynamics. Reinforcement learning (RL) is the representative example, demonstrating notable success in controlling legged robots in real-world applications \cite{gmm2021song,haarnoja2018soft, zhaoabsolute, lee2020learning, he2024learning, fu2024humanplushumanoidshadowingimitation, luo2023perpetual}. Despite their strong performance, model-free control laws are often highly task-specific and require retraining or fine-tuning when adapting to new tasks. 

Model-based approaches decouple the modeling of system dynamics from the realization of task objectives, enabling more efficient adaptation to new tasks. 
The models used in these approaches can be catagorized as: original nonlinear models, locally linearized models, and globally linearized models, with increasing modeling complexity and decreasing control complexity. 
For direct control of nonlinear models, \cite{yi2024covompctheoreticalanalysissamplingbased} introduced sampling-based MPC for legged robots, while \cite{Kazemi_Majd_Moghaddam_2013} applied robust backstepping control to quadruped robots. Local linearization methods, iLQR and NMPC often combine analytical \cite{sotaro2023analytical,zhu2024convergentilqrsafetrajectory,zhang2024robotsattitudesingularityfreequaternionbased} or learned models \cite{nagabandi2017neuralnetworkdynamicsmodelbased, liu2023modelbasedcontrolsparseneural} with complex controllers, requiring extensive domain expertise. A promising alternative to these approaches is employing Koopman Operator Theory to construct globally linearized models of legged systems and enabling the use of a simpler controller, e.g., linear MPC. Although Koopman-based methods have proven effective in motion planning \cite{kim2024learning} and gait prediction \cite{KROLICKI2022420}, no prior work has directly modeled the whole body dynamics of legged robots with Koopman Operator.

A Koopman Operator can be derived analytically or approximated through data-driven approaches. While analytical approaches are extremely challenging for high-dimensional nonlinear systems like legged robots, our method adopts a data-driven approach that reduces reliance on domain expertise and enhances generalizability. Recent works have applied data-driven approaches to obtain linear models for high-level motion planning \cite{kim2024learning, lyu2023task}. However, these methods typically depend on end-to-end optimization aimed at specific task objectives, which can limit their broader applicability. In contrast, \cite{shi2022deep, korda2018linear,mamakoukas2021derivative} have focused on obtaining accurate Koopman approximations for low-level control, but their applicability is limited to low-dimensional systems \cite{shi2024koopmanoperatorsrobotlearning}. 
It remains challenging to apply Koopman Operator to linearize high-dimensional nonlinear systems due to (a) imperfect approximations and (b) domain shifts arising from limited state transition data that only covers a subspace of the entire state space. Such approximation errors make the system sensitive to inaccurately modeled dynamics, while domain shifts cause failures under perturbations.

To address these challenges, we propose an incremental Koopman algorithm for high-dimensional legged robots that continually learn and lift Koopman dynamics. Our approach gradually expands both the training dataset and the latent space dimension.
The core idea is that this expansion progressively spans a state space where refined Koopman dynamics reduce linear approximation errors, ultimately converging to the true Koopman operator, as supported by our theoretical analysis. 
Experimental results demonstrate that our approach reduces linearization error within a few iterations and enables effective locomotion control for legged robots using Model Predictive Control (MPC). Our key contributions are summarized as follows:
\begin{enumerate}[leftmargin=12pt,itemsep=-5pt,  topsep=-10pt]
    \item We propose an iterative continual learning algorithm to mitigate linear approximation errors for Koopman dynamics, supported by a theoretical guarantee of convergence. 
    \item Our approach is the first work to implement locomotion control for high-dimensional legged robots using linearized whole body dynamics with Koopman Operator.
    \item Compared to traditional NMPC or RL mehods, our approach offers a scalable way to accurately model the dynamics of high-dimensional systems without requiring extensive expert knowledge, while maintaining generalizability across tasks and systems.
\end{enumerate}


\section{Problem Formulation}
\label{sec: problem formulation}

\subsection{Preliminary: Koopman Operator Theory}
\label{sec: koopman operator theo}
Koopman Operator Theory offers a powerful framework for analyzing nonlinear dynamical systems by mapping them into equivalent linear systems in an infinite-dimensional latent space. Consider the discrete-time nonlinear autonomous system $ s_{t+1} = f(s_t)$, where  $s_t \in \mathcal{S}$ represents the state. Koopman operator theory introduces the embedding function $\phi: \mathcal{S} \rightarrow \mathcal{O}$ to map the original state space $\mathcal{S}$ to an infinite dimensional latent space $\mathcal{O}$. In this latent space, the dynamics become linear through the Koopman Operator $\mathcal{K}:$
\begin{align}
\label{eq:autonomous_koopman}
    \mathcal{K} \phi(s) = \phi(f(s))
\end{align}
Consider a non-autonomous dynamical system with control input $u_t \in \mathcal{U} \subset \mathbb{R}^{m'}$ and system state $x_t \in \mathcal{X} \subset \mathbb{R}^{n'}$, given by $x_{t+1} = f(x_t, u_t)$. The Koopman Operator transforms the system dynamics into a linear form.
\begin{align}
    \label{eq: x u evolution}
    \mathcal{K}\phi(x_t, u_t) = \phi(f(x_{t}, u_{t})) = \phi(x_{t+1})
\end{align}

In practice, we approximate the Koopman operator by constraining the latent space to a finite-dimensional vector space. 
For a non-autonomous system, a common strategy is to define the embedding function as $\phi(x_t, u_t) = \left[ g(x_t); u_t \right]$, where $g: \mathbb{R}^{n'} \rightarrow \mathbb{R}^n$ is a state embedding function. The Koopman operator $\mathcal{K}$ can be approximated by a matrix representation $K = \left[\begin{matrix} A\in\mathbb{R}^{n\times n} & B\in\mathbb{R}^{n\times m'} \\ C\in\mathbb{R}^{m'\times n} &D\in\mathbb{R}^{m' \times m'} \end{matrix}\right]$. In conjunction with \eqref{eq: x u evolution},  we obtain a linear model in the lifted space with the state evolution:
\begin{align}
    \label{eq: Ax + Bu}
    g(x_{t+1}) = Ag(x_t) + Bu_t
\end{align}

To preserve state information with clear semantics \cite{shi2022deep}, and avoid the case of degeneration, such as cases where $A=B=0 \text{{~and~}} g(x)\equiv 0$, we concatenate the original state with the neural network embedding to define the latent state $z_t$ as $z_t = g(x_t) = \left[ \begin{matrix} x_t, g'(x_t) \end{matrix} \right]^\top$, where $g': \mathbb{R}^{n'} \rightarrow \mathbb{R}^{n-n'}$ is a parameterized neural network. 
This approach allows us to retrieve the original state using a linear transformation.
\begin{align}
    x_{t+1} = P z_{t+1} =P g(x_{t+1}) , ~ P=\left[I_{n'\times n'}, \textbf{0}_{n'\times (n-n')}\right] \in \mathbb{R}^{n' \times n}
\end{align}

This formulation supports solving state-constrained control problems (e.g., collision avoidance) by directly transforming the constraints on $x$ to constraints on $z$ without compromising the lifted system's linear properties.
Additionally, we use an end-to-end training approach for both the embedding function and the Koopman operator,  parameterized as $\mathcal{T} \doteq (g, A, B)$. 

\subsection{Linear Model Predictive Control}
\label{sec: mpc}


With the linear model from the Koopman operator, in our high-dimensional legged robot control tasks, we employ a linear MPC algorithm as the controller $\pi_{mpc}$, thus the problem can be easily solved by Quadratic Programming (QP). The optimization formulation at time step $t$ is as follows:
\begin{align}
    \label{eq: mpc formulation}
    \underset{u_{t:t+H-1}}{\min} & ~~\| Pz_{t:t+H-1} - x^*_{t:t+H-1} \|^2_Q + \| u_{t:t+H-1} \|^2_R 
    + \| Pz_{t+H} - x^*_{t+H} \|^2_F \\ \nonumber
    \textit{s.t.}& ~~ z_{t+k+1} = Az_{t+k} + Bu_{t+k}, 
     ~~ u_{t+k} \in \left[ u_{min}, u_{max} \right], ~~~\forall k = 0, \cdots, H-1
\end{align}
where $H$ denotes the horizon length, $x^*_{t:t+H}$ represents the reference trajectory, $Q$, $R$, $F$ are cost matrices and $\| x \|^2_Q$ is defined as $x^TQx$. The range $\left[ u_{min}, u_{max} \right]$ specifies the valid interval for control inputs. The first constraint ensures adherence to the learned dynamics in the latent space, while the second constraint bounds control signals within the valid region.
Unlike model-free methods, which may only require a planar velocity objective, linear MPC relies on a full-body reference trajectory. Developing methods to track lower-dimensional references is left for future work.

\subsection{Legged Robot System}
\label{sec:problem_legged_robot}
As demonstrated in \cite{Wieber2016ModelingAC}, legged robot systems inherently possess the following characteristics: (i) discontinuities resulting from mode transitions, (ii) sensitivity to noisy control inputs while maintaining balance, and (iii) failure situations that are often irrecoverable. Existing Koopman-based methods focus mainly on low-dimensional continuous systems with fixed contact modes \cite{shi2022deep}, where the dynamics are continuous and easier to fit. However, when applied to high-dimensional hybrid legged systems, the complex and diverse contact modes introduce significant discontinuities, greatly complicating the modeling process. Moreover, these existing Koopman methods, which exhibit considerable approximation errors and limited latent subspace \cite{han2023utility}, render hybrid legged systems prone to irrecoverable failures caused by unmodeled dynamics and external perturbations. To overcome these challenges, we  introduce an algorithm that accurately and robustly models dynamics through continual learning and lifting, effectively expanding a resilient latent subspace.

\section{Incremental Koopman Algorithm}
\label{sec: incremental koopman}

\begin{figure}[t]
    \centering
    \includegraphics[width=0.99\textwidth]{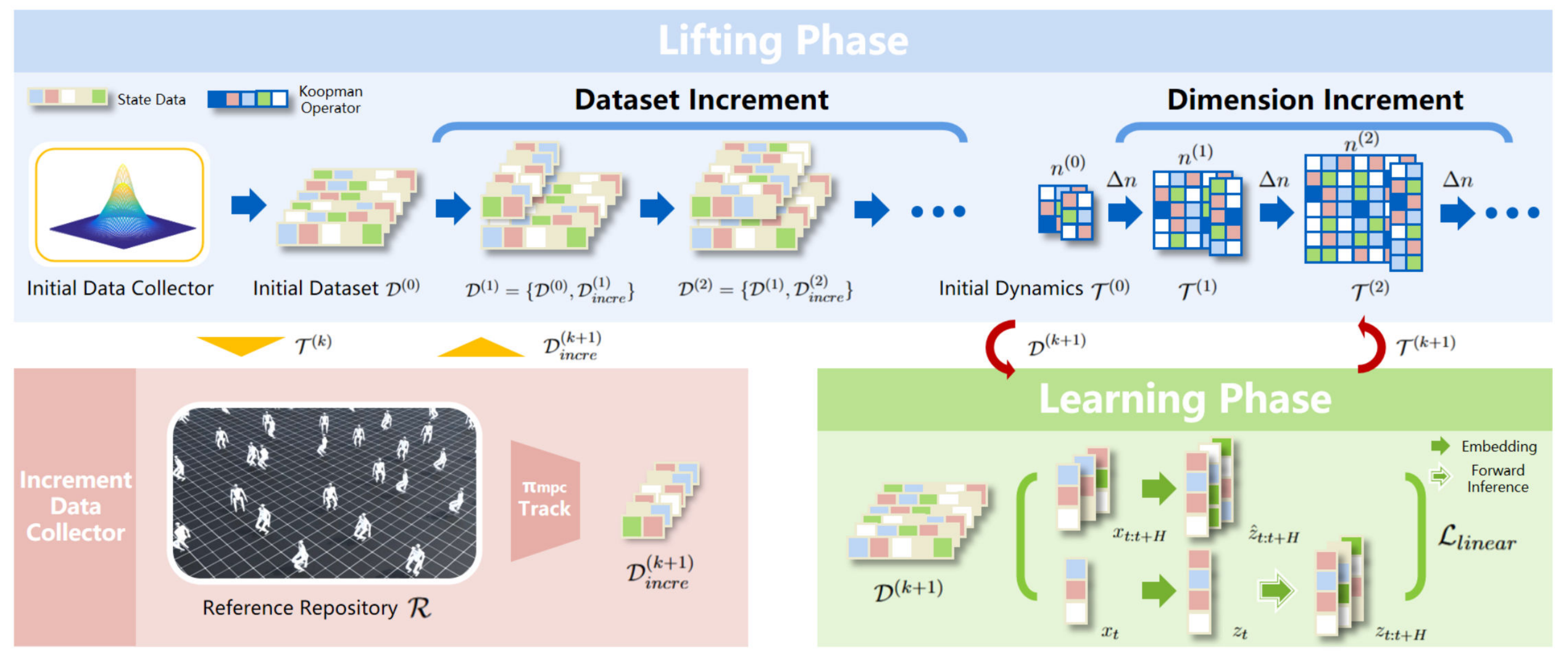}
    \vspace{-5pt}
    \caption{Overview of Incremental Koopman Algorithm}
    \label{fig: overview}
    \vspace{-10pt}
\end{figure}

As shown in \Cref{fig: overview}, the pipeline consists of two data collectors and two phases. After the initial data collection and dynamics learning, we iteratively alternate between the lifting phase, where the dataset is expanded by the increment data collector and the latent space dimension is increased, and the learning phase, during which dynamics are retrained based on the updated dataset and dimension.

\subsection{Initialization and Increment of Dataset}
\label{sec: init and incre of data}

\textbf{~~~~~~Initial Data Collector.}
Unlike low-dimensional continuous systems explored in \cite{shi2022deep}, the high-dimensional hybrid dynamics of legged robots need training data with reasonable gait and consistent contact modes to form a meaningful latent subspace. Thus, we use an initial data collector (e.g., an RL policy or tele-operation) to generate dynamically feasible trajectories with reasonable movement patterns for the construction of the initial dataset $\mathcal{D}^{(0)}$.

\textbf{Increment Data Collector.}
Our method enables robust tracking within the reference repository $\mathcal{R}$, where references may be dynamically feasible or infeasible, with gaits and contact modes similar to $\mathcal{D}^{(0)}$. While $\mathcal{D}^{(0)}$ lays the foundation for the latent subspace, learning from limited samples—especially from a “good behavioral collector”—is insufficient. Models may still exhibit significant approximation errors in near-failure or failure scenarios, leading to ineffective tracking. Thus we use the MPC controller in \Cref{sec: mpc} to track references in $\mathcal{R}$, collecting failed tracking data $\mathcal{D}_{incre}$ to naturally expand the dataset, thereby enhancing the robustness of the latent subspace. 

Notably, our algorithm can be applied to any eligible initial data collector and reference repository $\mathcal{R}$. In this work, within a deterministic environment, we use a Proximal Policy Optimization (PPO \cite{schulman2017proximalpolicyoptimizationalgorithms}) policy to build $\mathcal{D}^{(0)}$ and the initial reference repository $\mathcal{R}_{init}$, then add uniform noise in the range [-0.05, 0.05] to all references, generating the dynamically infeasible $\mathcal{R}$. 

\subsection{Training with Given Data}
\label{sec: training with data}
Based on the base Koopman framework introduced in \Cref{sec: koopman operator theo}, our primary goal is to learn the embedding function $g$ and the Koopman operators $A, B$ end-to-end. To achieve this, we employ a discounted k-step prediction loss to enhance long-horizon prediction capabilities which is crucial for MPC control. Given a set of trajectory data $\{x_{t:t+H}; u_{t:t+H-1}\}$, real latent trajectories can be computed as $\{z_{t:t+H}=g(x_{t:t+H})\}$. At the same time, predicted latent trajectories can be obtained through inference: $\{\hat{z}_{t:t+H} | \hat{z}_t = z_t, \hat{z}_{t+h} = A\hat{z}_{t+h-1}+Bu_{t+h-1}, h=1,2,\cdots,H\}$. Thus the loss function is defined as:
\begin{align}
    \label{eq: loss}
    \mathcal{L}_{koopman} = \frac{1}{H}\sum_{h=1}^H \gamma^h\left(  \underbrace{\| \hat{z}_{t+h} - z_{t+h}\|^2}_{\mathcal{L}_{linear}} + \alpha\cdot \underbrace{\|\hat{x}_{t+h} - x_{t+h}}_{\mathcal{L}_{recon}} \|^2\right)
\end{align}
where $\hat{x}_{t:t+H} = P\hat{z}_{t:t+H}$, $\alpha$ is the weight of $\mathcal{L}_{recon}$ and $\gamma$ is the discount factor. Here $\mathcal{L}_{linear}$ focuses on the linearization effect, while $\mathcal{L}_{recon}$ addresses the reconstruction effect. Since $z_t = [x_t, g'(x_t)]^T$, $\mathcal{L}_{linear}$ inherently implies $\mathcal{L}_{recon}$. Thus, slighty emphasizing reconsrtuction of original state by setting $\alpha$ to 0.1 shows better performance in practice. Then we can derive dynamics $\mathcal{T}^{(k)} \doteq (g^{(k)}_{n^{(k)}}, A^{(k)}_{n^{(k)}}, B^{(k)}_{n^{(k)}})$ for dataset $\mathcal{D}^{(k)}$, here $n^{(k)}$ represents the latent space dimension of iteration $k$. 

\subsection{Continual Learning and Lifting}
The incremental Koopman algorithm begins with the initial dynamics $\mathcal{T}^{(0)}$ trained on the initial dataset $\mathcal{D}^{(0)}$. During the lifting phase, we use the increment data collector introduced in \Cref{sec: init and incre of data} to gather the incremental dataset $\mathcal{D}^{(1)}_{incre}$, which is then used to augment the initial dataset, forming $\mathcal{D}^{(1)} = \mathcal{D}^{(0)} \cup \mathcal{D}^{(1)}_{incre} $. Simultaneously, to accommodate the growing dataset, the latent space dimension is updated as $n^{(1)} = n^{(0)} + \Delta n$, where $\Delta n$ is a hyperparameter chosen based on the trade-off between computational complexity, performance gain, and the requirement $m = \Omega(n \log(n))$ to be introduced in \Cref{sec: theoretical analysis}, where $m$ denotes the dataset size. 

In the learning phase, we utilize the loss introduced in \Cref{sec: training with data} to obtain 
$\mathcal{T}^{(1)} = (g^{(1)}_{n^{(1)}}, A^{(1)}_{n^{(1)}}, \\B^{(1)}_{n^{(1)}})$. 
This process is repeated until the survival steps $T_{sur}$ under MPC control with dynamics $\mathcal{T}^{(k)}$ cease to show improvement over $\mathcal{T}^{(k-1)}$. The details are shown in \Cref{alg:koopman}.
While sampling all state-action pairs is unrealistic and unscalable in high-dimensional problems, our method balances sample complexity, approximation quality, and scalability. It leverages on-policy exploration to gradually minimize approximation errors within the MPC controller's region of attraction, enabling reliable decisions and establishing robust linear latent dynamics. We validate our claims in \Cref{sec: exp}.

\subsection{Theoretical analysis}
\label{sec: theoretical analysis}

Incremental increases in the dimension of latent space and data size are central to our proposed algorithm. This section demonstrates that, under our incremental strategy, the learned Koopman operator matrix  $K$ converges to the true Koopman operator $\mathcal{K}$. For theoretical analysis, we primarily focus on autonomous systems as described in \eqref{eq:autonomous_koopman}. For non-autonomous systems, the methodology extends by setting $s=[x;u]^\top$. 
The fundamental premise, as established in \cite{korda2018convergence, korda2018linear}, is that the learned Koopman operator $K$ is the $L_2$ projection of the true
Koopman operator $\mathcal{K}$ onto the span of the $n$-dimensional embedding functions $\phi(s)=[\phi_1(s), \cdot, \phi_n(s)]$, where $\phi_i(s)$ is the one-dimensional embedding function for the $i$-th dimension of the latent space. Unlike the convergence analysis in \cite{korda2018convergence}, our theoretical framework introduces the convergence rate of our incremental Koopman algorithm. The detailed proof can be found in \cref{sec:sup_proof}.

\textit{
\textbf{Theorem 1.} Under the assumptions of 1) data samples $s_1, \cdot, s_m$ are i.i.d distributed; 2) the latent state is bounded, $\|\phi(s)\| < \infty$; 3) the embedding functions $\phi_1, \cdots, \phi_n$ are orthogonal (independent).\\
(a) The learned Koopman operator converges to the true Koopman Operator:
\begin{align}
    \lim_{m=\Omega(n ln(n)), n \rightarrow \infty} K \rightarrow \mathcal{K}.
\end{align} 
(b) Assuming additional conditions: 4) The eigenvalues of $\mathcal{K}$ decay sufficiently fast, i.e., $ | \lambda_i | \leq \frac{C}{i} $ for some constant $ C > 0 $. 5) The embedding functions $\phi(\cdot) = [\phi_1, \cdots,\phi_n]^\top$ correspond to the first $n$ eigenfunctions of $\mathcal{K}$ associated with the largest eigenvalues in magnitude.
Then the convergence rate of the linear approximation error is given by:
\begin{align}
 \text{error} \leq \mathcal{O}( \sqrt{ \frac{\ln(n)}{m} })+  \mathcal{O}\left( \frac{1}{\sqrt{n}} \right).
\end{align} 
}

As indicated in Theorem 1, the proposed algorithm, by incrementally increasing the latent dimension $n$ and data size $m$, progressively reduces the linear approximation error. According to Theorem 1(b), if the sample size meets $m=\Omega(n \ln(n) )$, this reduction error follows the rate of $ \mathcal{O}(n^{-\frac{1}{2}})$ as the dimension increases. 
In our experiments, we adhere to this guideline to ensure adequate sample generation. We note that the above assumptions are generally mild and hold in many practical scenarios.  Further details on the theorem and underlying assumptions can be found in \cref{sec:sup_proof}.


\section{Experiments}
\label{sec: exp}

In our experiments, we evaluate the proposed method by addressing the following questions:
\textbf{Q1} Does the learned Koopman Dynamics from \cref{alg:koopman} achive better state prediction performance compared to baselines?
\textbf{Q2} Does our approach, combining learned Koopman dynamics with MPC control, achieve superior tracking performance relative to baselines?
\textbf{Q3} Is the continual expansion of the dataset size in our proposed method effective? 
\textbf{Q4} Is the continual increase in the dimensions of the latent state beneficial?

\subsection{Experiment Setup}
\label{sec: exp setup}

\textbf{~~~~~~Task settings.} 
Our experiments use IsaacLab \cite{mittal2023orbit} as the simulation platform, the most advanced simulator offering comprehensive and flexible support for a wide range of robots and environments. Five different legged robots are included in our experiments: (i) \textbf{ANYmal-D: } (\Cref{fig:anymal-d}) A quadruped robot ($\mathcal{U} \subseteq \mathbb{R}^{12}$) designed by ANYbotics. (ii) \textbf{Unitree-A1: } (\Cref{fig:a1}) A quadruped robot ($\mathcal{U} \subseteq \mathbb{R}^{12}$) designed by Unitree. (iii) \textbf{Unitree-Go2: } (\Cref{fig:a1}) A quadruped robot ($\mathcal{U} \subseteq \mathbb{R}^{12}$) designed by Unitree.  (iiiv) \textbf{Unitree-H1: } (\Cref{fig:a1}) A Humanoid ($\mathcal{U} \subseteq \mathbb{R}^{19}$) designed by Unitree.  (v) \textbf{Unitree-G1: } (\Cref{fig:a1}) A Humanoid ($\mathcal{U} \subseteq \mathbb{R}^{23}$) designed by Unitree. 
Furthermore, two different types of terrain are considered, including (i) \textbf{Flat: } (\Cref{fig:flat}) flat terrain. (ii) \textbf{Rough: } (\Cref{fig:rough}) rough terrain with random surface irregularities, where the height follows a uniform distribution between 0.005 and 0.025, and the minimum height variation is 0.005 (in m).
All test suites are based on the task \textbf{Walk}: tracking a velocity command with uniformly sampled heading direction, x-axis linear velocity, and y-axis linear velocity. 
Considering these settings, we design 7 test suites with 5 types of legged robots and 2 types of terrain, which are summarized in \Cref{tab: test suites}. We name these test suites as \texttt{\{Terrain\}-\{Make\}-\{Model\}}.

\textbf{Comparison Group.}
We compare Incremental Koopman algorithm with state-of-the-art model-based control algorithms (i) Deep KoopmanU with Control (DKUC, \cite{shi2022deep}) algorithm, (ii) Deep Koopman Affine with control (DKAC, \cite{shi2022deep}) algorithm, (iii) Neural Network Dynamics Model (NNDM, \cite{nagabandi2017neuralnetworkdynamicsmodelbased, liu2023modelbasedcontrolsparseneural}) with Nonlinear MPC (iv) Deep Koopman Reinforcement Learning (DKRL, \cite{gmm2021song}). It is worth noting that all methods are trained seperately on 7 test suites. And for all experiments, we take the best algorithm-specific parameters mentioned in the original paper and keep the common parameters the same.

\textbf{Metrics.} 
We introduce the following metrics to evaluate the tracking ability of our algorithm: (i) \textbf{Prediction-based: } we leverage k-step prediction error $E_{pre}(k) \doteq \frac{1}{kn'}\sum_{t=1}^k \| x_t - x_t^* \|_1 ~(x_0 = x_0^*)$ generally used for measuring prediction ability of given dynamics. (ii) \textbf{Pose-based: } To evaluate tracking accuracy, we evaluate Joint-relative mean per-joint position error ($E_{JrPE}$), Joint-relative mean per-joint velocity error ($E_{JrVE}$), Joint-relative mean per-joint acceleration error ($E_{JrAE}$), Root mean position error ($E_{RPE}$), Root mean orientation error ($E_{ROE}$), Root mean linear velocity error ($E_{RLVE}$) and Root mean angular velocity error ($E_{RLAE}$). Check \Cref{sec: def of metrics} for detailed equations.  (iii) \textbf{Physics-based: } Considering the easy-to-fail feature of legged robots, for the same reference repository $\mathcal{R}$, we evaluate the average survival simulation time step $T_{Sur}$ (set 200 as upper bound, simulation frequency is 50Hz) for each algorithm, deeming tracking unsuccessful when $E_{JrPE}$ is larger than $\epsilon_{fail}$. Check \Cref{tab: test suites} for details.

\textbf{State.} 
The system state $x_t \doteq \left\{ \begin{matrix} (j_t, \dot{j_t}, p^z_t, \dot{p_t}, \dot{r_t}), \textit{~for humanoid}~~~~~~~~~~~~ \\ (j_t, \dot{j_t}, p^z_t, \dot{p_t}, r_t, \dot{r_t}), \textit{~for quadruped robot} \end{matrix} \right.$ of dynamics and controller includes the joint position $j_t$ and joint velocity $\dot{j_t}$ in local coordinates, root height $p^z_t$, 3D root linear velocity $p_t$, 4D root orientation $r_t$ (optional for humanoid and represented as a quaternion), and 3D root angular velocity $\dot{r_t}$ in world coordinates. 
All of the aforementioned quantities are normalized to follow a standard Gaussian distribution, $\mathcal{N}(0,1)$, in order to eliminate inconsistencies in the scale of different physical quantities. 

\textbf{Control Input and Low-level Controller.}
We use a proportional-derivative (PD) controller at each joint of the legged robots as the low-level controller. Thus, the control input $u_t$ determined by the MPC serves as the joint target, denoted as $j_t^d = u_t$. The torque applied to each joint is given by $\tau_t = k^p \circ (j_t^d - j_t) - k^d \circ \dot{j_t}$, where $k^p$ and $k^d$ are the proportional and derivative gains, respectively. The low-level controller operates at a frequency of 200 Hz, with a decimation factor of 4.

\subsection{K-step Prediction Error}

\begin{figure}[t]
    \centering
    \includegraphics[width=0.99\textwidth]{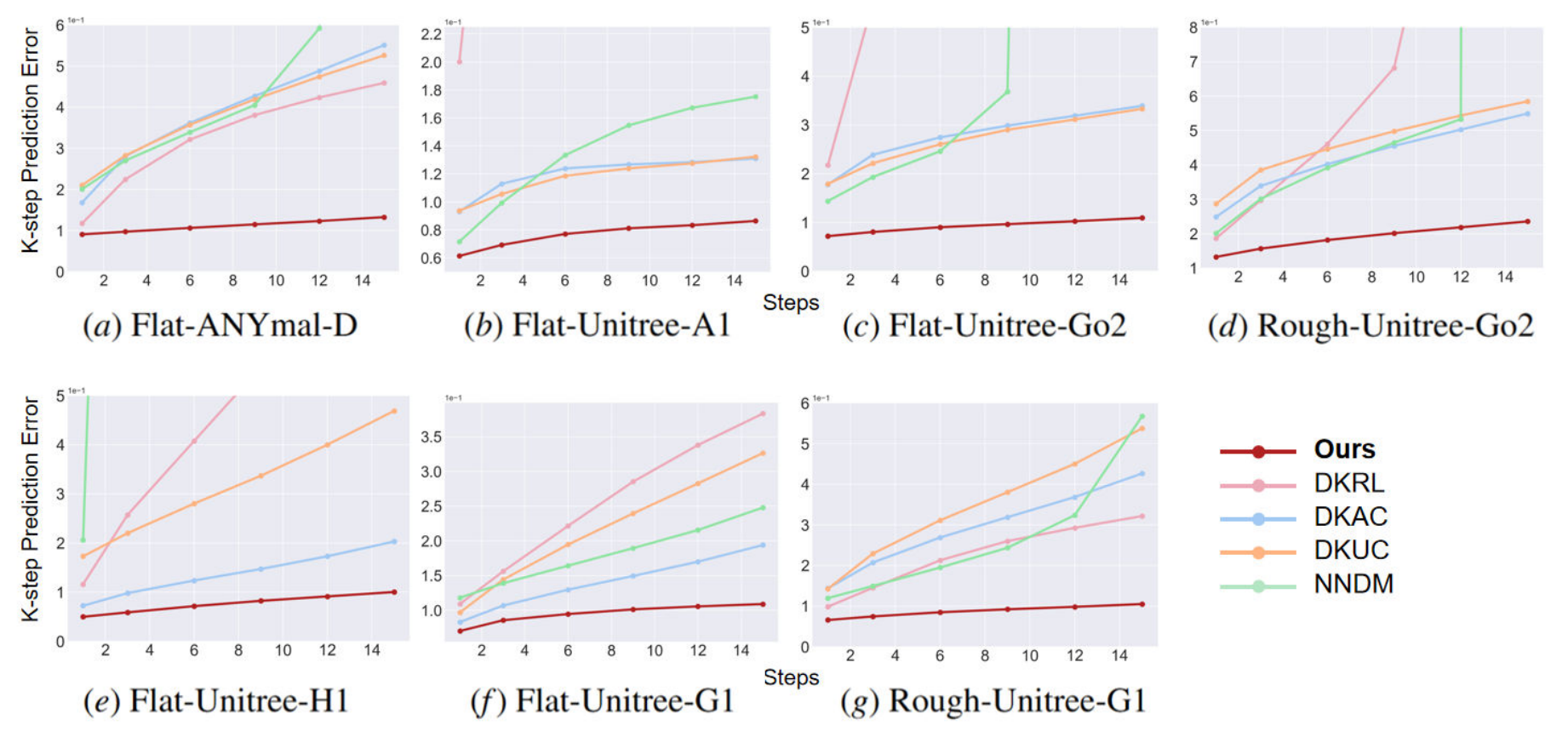}
    \vspace{-8pt}
    \caption{Comparison results of k-step prediction error in our test suites.}
    \label{fig: k step predict}
    \vspace{-10pt}
\end{figure}

Specifically, we set $k$ to $[1, 3, 6, 9, 12, 15]$ and compute $E_{pre}(k)$ for each algorithm, subsequently plotting the results in \Cref{fig: k step predict}. 
The results highlight our algorithm's exceptional ability to maintain low prediction errors, even over long horizons in complex, nonlinear tasks—challenging for other algorithms in our comparison. 
Figures \textcolor{blue}{3(c)} to \textcolor{blue}{3(e)} show that NNDM and DKRL struggle with the accumulation of explosive errors as horizon length $k$ increases, while our algorithm remains stable, with negligible error growth.
Though DKAC and DKUC manage a stable rate of loss increase, they exhibit relatively high prediction errors due to their limited subspace modeling compared to our incremental method. Please check \Cref{sec: dataset settings} for details of test dataset. These results address \textbf{Q1}. 

\subsection{Tracking Performance}
\label{sec: tracking}

\begin{figure}[h]
    \centering
    \includegraphics[width=0.99\textwidth]{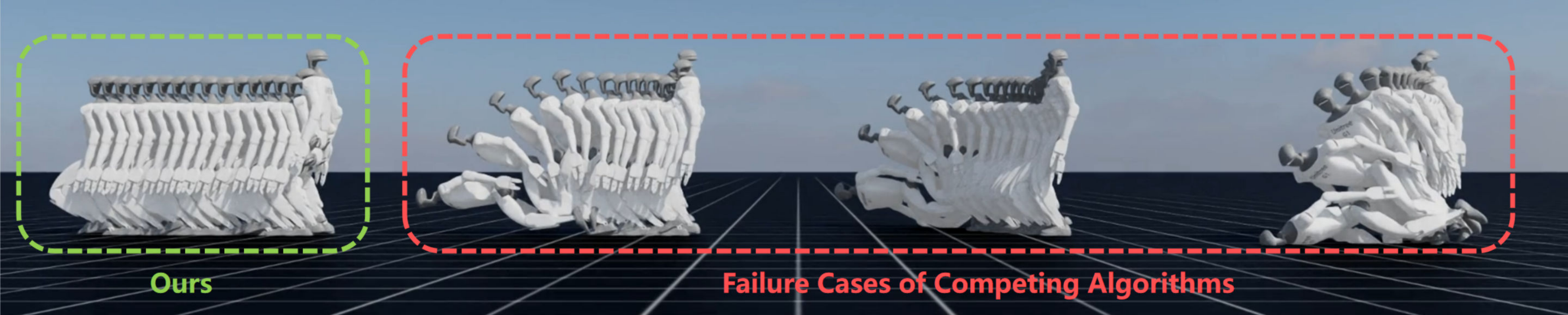}
    \caption{Visualization of tracking performance on Flat-Unitree-G1.}
    \label{fig: failure cases}
    \vspace{-10pt}
\end{figure}

\begin{table*}[h]
\begin{center}
\resizebox{0.95\textwidth}{!}{
    \begin{tabular}{c|ccc|cccc|c}
    \toprule
    \multicolumn{9}{c}{Synthesised Tracking Metrics}\\
    \hline \\[-0.95em]
    \multicolumn{1}{c|}{Algorithm}&\multicolumn{3}{c}{Joint-relative $\downarrow$} & \multicolumn{4}{|c|}{Root-relative $\downarrow$}& \multicolumn{1}{c}{Survival $\uparrow$}\\
    \cline{2-9}\\[-1.02em]
     & $E_{JrPE}$ & $E_{JrVE}$ & $E_{JrAE}$ & $E_{RPE}$ & $E_{ROE}$ & $E_{RLVE}$ & $E_{RAVE}$ & $T_{Sur}$ \\
    \hline \\[-0.95em]
    \textbf{Ours} with MPC& \textbf{0.0348} & \textbf{0.6499} & \textbf{43.1465} & \textbf{0.1231} & \textbf{0.0668} & \textbf{0.1216} & \textbf{0.3289} & \textbf{188.4514} \\
    DKRL with MPC & 0.0823 & 1.1251 & 68.9520 & 0.2978 & 0.1561 & 0.2089 & 0.5634 & 116.9540 \\
    DKAC with MPC & 0.1816 & 2.0694 & 117.5515 & 0.3955 & 0.2749 & 0.2888 & 0.9143 & 25.0254 \\
    DKUC with MPC & 0.1576 & 1.0828 & 50.5745 & 0.2934 & 0.1989 & 0.2252 & 0.5559 & 82.4620 \\
    NNDM with NMPC & 0.1439 & 2.2020 & 127.4524 & 0.4334 & 0.2506 & 0.2996 & 0.8536 & 35.4709 \\
    \bottomrule
    \end{tabular}
}
\caption{The average tracking metrics evaluated for each algorithm across all 7 test suites.}
\label{tab: tracking performance}
\end{center}
\vspace{-20pt}
\end{table*}

To highlight the advantage of our method in tracking tasks, we track 3000 references in $\mathcal{R}$ over 200 steps in parallel, presenting the results in \Cref{tab: tracking performance} with average tracking metrics from \Cref{sec: exp setup}. Further details are provided in \Cref{sec: total tracking exp}.

Compared to other algorithms, our approach shows (i) high $T_{Sur}$ values nearing the upper bound of 200, (ii) precise, continuous joint-level tracking, and (iii) sustained global tracking with balance and stability. All measured metrics are significantly lower for competing algorithms, which often fail to track correctly and fall after only a few attempts. In contrast, our algorithm delivers efficient, accurate tracking with a steady gait, demonstrating its ability to infer corner cases during tracking, thanks to the continual learning and lifting process in \Cref{alg:koopman}. Meanwhile, others struggle to achieve the correct poses and gait when faced with falling. Notably, the 200-step upper bound for $T_{Sur}$ suggests that many test cases remain robust enough for even longer tracking durations.

Although DKRL and DKUC achieve relatively high $T_{Sur}$, their $E_{JrPE}$ and $E_{RPE}$ are 2 to 5 times higher than ours, indicating that their metrics include brief failure episodes. In contrast, our algorithm maintains robust tracking throughout, showcasing superior steady tracking with minimal error. DKAC and NNDM, however, suffer from short inference capabilities and sensitivity to the dynamics' corner cases, making them prone to failure—especially in situations where recovery is impossible for legged robots. As shown in \Cref{fig: failure cases}, unrecoverable failures end the tracking process, answering \textbf{Q2}.


\begin{figure}[t]
    \centering
    \includegraphics[width=0.99\textwidth]{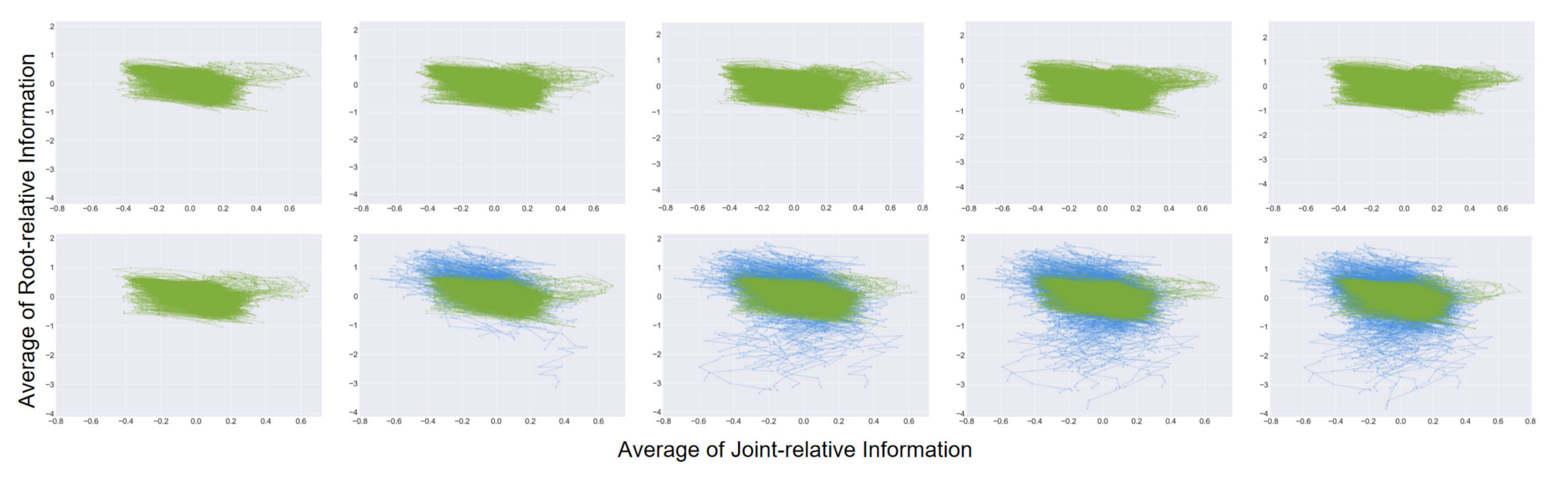}
    \vspace{-5pt}
    \caption{Visualization of the data distribution by plotting the means of the joint-relative and root-relative states. The first row illustrates the dataset expansion achieved using an RL policy, while the second row shows the results from applying our algorithm for dataset expansion.}
    \label{fig: evolution process of data}
    \vspace{-15pt}
\end{figure}

\subsection{Ablation on Dataset Increment}
\label{sec: ablation on data incre}
\begin{wrapfigure}{hbtp}{0.6\textwidth}
    \vspace{-20pt}
    \centering
    \subfigure[\small{Flat-Unitree-Go2}]{
        \centering
        \includegraphics[width=0.275\textwidth]{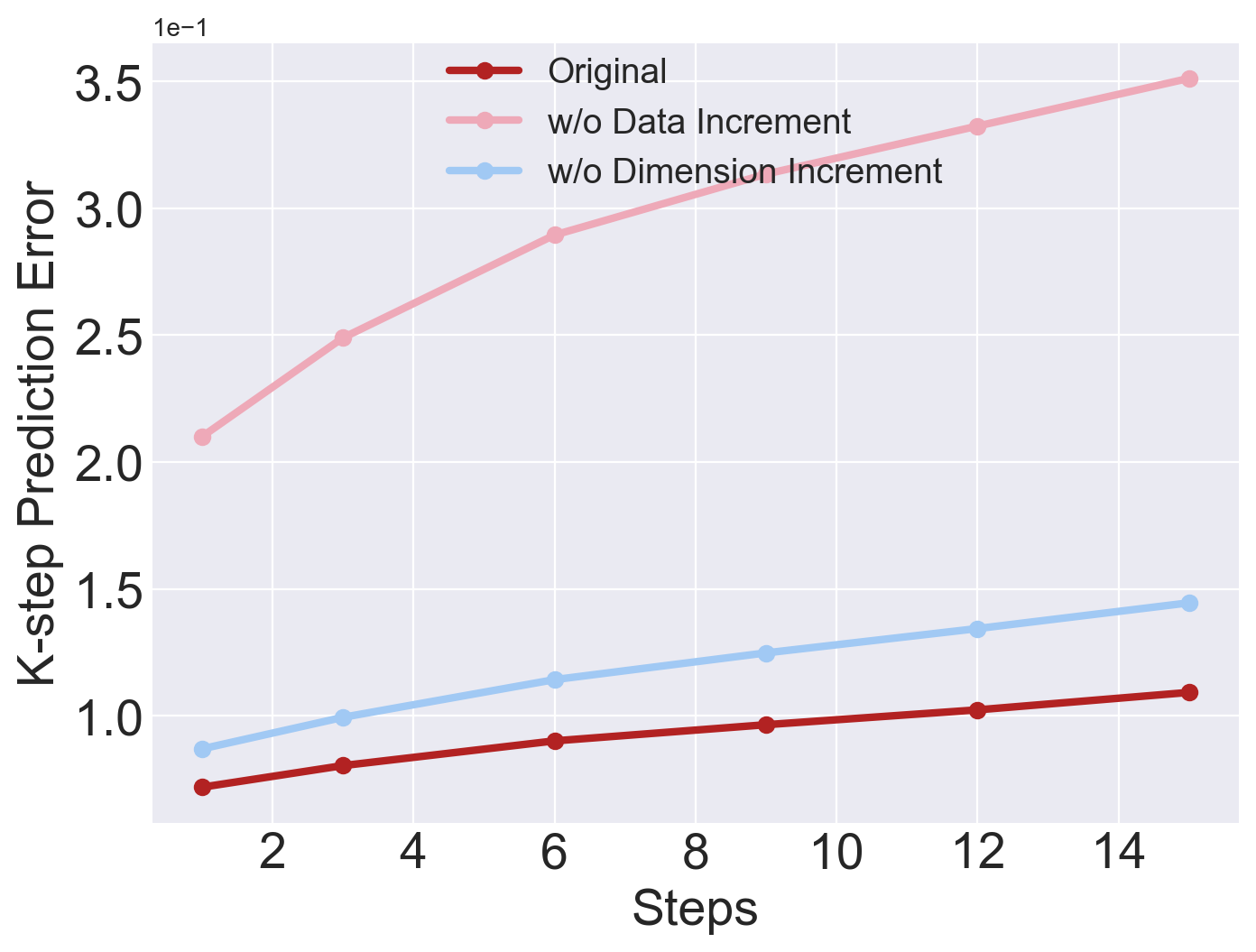}
        \label{fig:go2 ablation}
    }
    \subfigure[\small{Flat-Unitree-G1}]{
        \centering
        \includegraphics[width=0.275\textwidth]{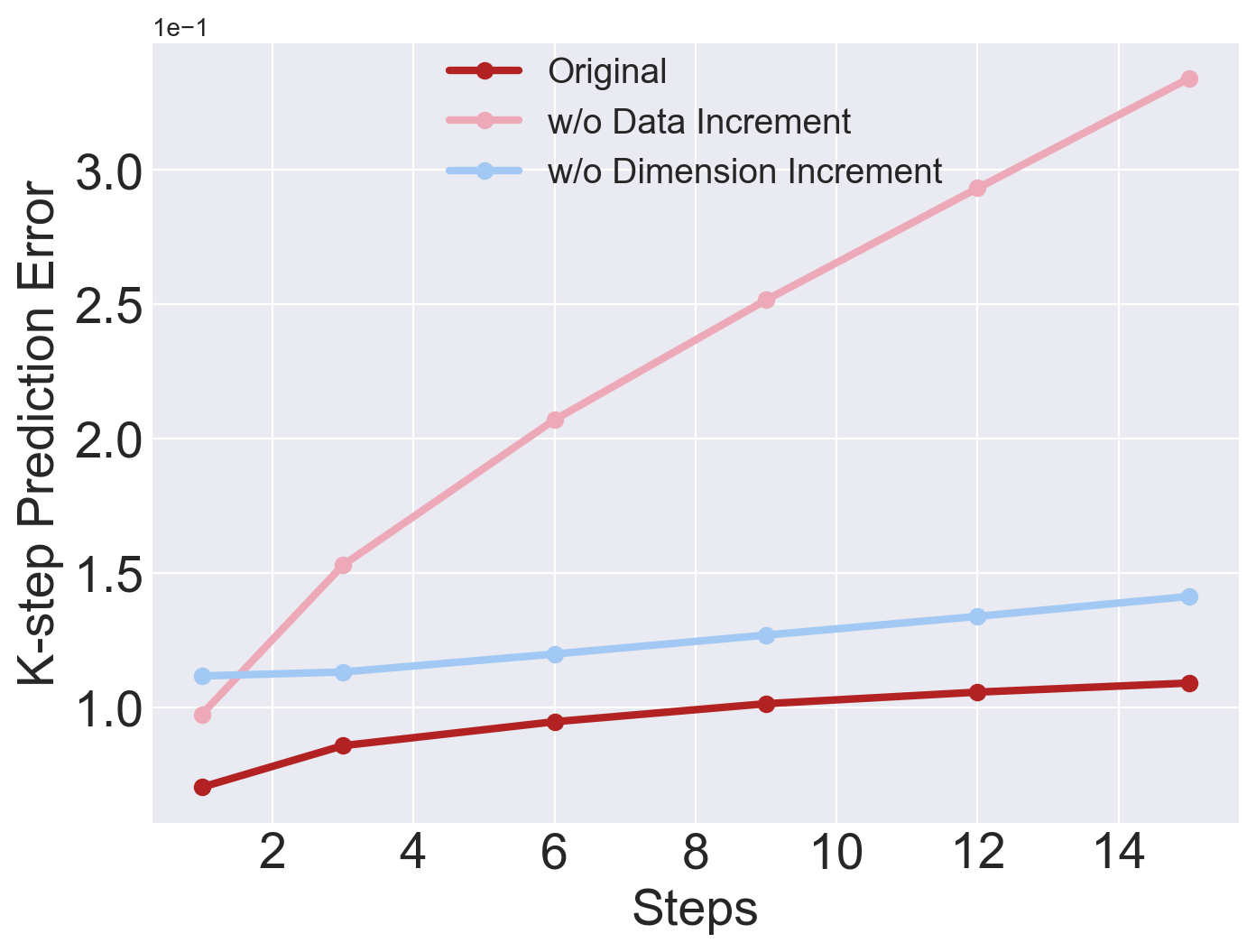}
        \label{fig:g1 ablation}
    }
    \vspace{-5pt}
    \caption{K-step prediction error comparison of ablation on dimension and dataset increment.} 
    \label{fig: ablation on increment}
    \vspace{-10pt}
\end{wrapfigure}

For the ablation study on dataset increment, we select Flat-Unitree-Go2 and Flat-Unitree-G1 to examine the impact of dataset increment techniques. As shown in \Cref{tab: tracking ablation,fig: ablation on increment}, the k-step prediction error rises sharply with increasing steps, highlighting limited inferencing capability when dynamics encounter corner cases due to insufficient latent subspace modeling. Furthermore, $E_{JrPE}$ is nearly seven times higher than the original method, indicating frequent failure scenarios. The low survival steps $T_{Sur}$ further corroborate this observation.
In contrast, the original algorithm \Cref{alg:koopman} maintains exceptional tracking performance and loss control, demonstrating the effectiveness of dataset enlargement. Additionally, \Cref{fig: evolution process of data} visualizes the data enlargement process, showing that RL policy-based expansion is ineffective due to repetitive data, while our algorithm achieves effective dataset growth. This underscores the importance of robust subspace modeling and accurate inferencing, addressing \textbf{Q3}.

\begin{table*}[t]
\begin{center}
\resizebox{0.95\textwidth}{!}{
    \begin{tabular}{c|ccc|cccc|c}
    \toprule
    \multicolumn{9}{c}{Average Tracking Metrics of Flat-Unitree-Go2 and Flat-Unitree-G1}\\
    \hline \\[-0.95em]
    \multicolumn{1}{c|}{Algorithm}&\multicolumn{3}{c}{Joint-relative $\downarrow$} & \multicolumn{4}{|c|}{Root-relative $\downarrow$}& \multicolumn{1}{c}{Survival $\uparrow$}\\
    \cline{2-9}\\[-1.02em]
     & $E_{JrPE}$ & $E_{JrVE}$ & $E_{JrAE}$ & $E_{RPE}$ & $E_{ROE}$ & $E_{RLVE}$ & $E_{RAVE}$ & $T_{Sur}$ \\
    \hline \\[-0.95em]
    Original & \textbf{0.0246} & \textbf{0.5954} & \textbf{42.1403} & \textbf{0.0673} & \textbf{0.0245} & \textbf{0.0650} & \textbf{0.2060} & \textbf{196.6190}\\
    w/o Data.I. & 0.2061 & 1.5251 & 65.5157 & 0.3452 & 0.2449 & 0.2740 & 0.6303 & 53.0540\\
    w/o Dim.I. & 0.1189 & 1.1743 & 63.6319 & 0.2526 & 0.1936 & 0.2178 & 0.5782 & 100.9350\\
    \bottomrule
    \end{tabular}
}
\caption{The average tracking metrics evaluated for the ablation studies. `w/o Data.I.' and `w/o Dim.I.' indicate the performance of the original method without data increment and dimension increment techniques, respectively}
\label{tab: tracking ablation}
\end{center}
\vspace{-25pt}
\end{table*}

\subsection{Ablation on Dimension Increment}

To conduct the ablation study on dimension increment, we select the same test suites mentioned in \Cref{sec: ablation on data incre}. As shown in \Cref{fig: ablation on increment}, the k-step prediction error remains steady and relatively low, similar to the original methods, benefiting from the continual increment of the dataset to model the subspace. However, as indicated in \Cref{tab: tracking ablation}, although the algorithm without dimension increment outperforms the algorithm without data increment, its tracking performance still falls significantly short of the original method. This suggests that a low latent space dimension fails to model robust dynamics strong enough to resist unseen noise during simulation. Furthermore, the lack of inferencing capability highlights its shortcomings in linearization and generalization. Thus, the dimension increment technique emphasizes the importance of modeling robust and general sub-dynamics, addressing \textbf{Q4}.

\section{Conclusion and Future Works}
By progressively expanding the dataset and latent subspace, our method ensures convergence of linearization errors and enables accurate approximations of the system dynamics. Experimental results show that the approach achieves high control performance with simple MPC controllers on various tasks after just a few iterations. This work represents the first successful application of linearized Koopman dynamics to locomotion control of legged robots, offering a scalable and model-based control solution. At the same time, the incremental Koopman algorithm may encounter an explosion in latent dimensionality when the subspace expands rapidly. In the future, tele-operation or retargeted human data will be tested to implement our algorithm in real-world scenarios.

\acks{
This research is supported by the National Science Foundation (NSF) under Grant No. 2144489. 
}


\bibliography{main.bib} 

\newpage
\appendix
\section{Related Works}
\label{sec: related works}
\textbf{Controlling Nonlinear Dynamics.}
Nonlinear dynamic systems are generally controlled using two main approaches: model-free and model-based methods.
Model-free methods, often associated with learning-based approaches, have become popular for addressing complex, high-dimensional systems in a data-driven manner. Reinforcement learning (RL) techniques, for instance, approximate a combination of system dynamics and control policies by training neural networks using methods like policy gradients or value iteration. Off-policy algorithms such as SAC \cite{haarnoja2018soft} efficiently reuse data to learn control policies, whereas on-policy methods like PPO \cite{schulman2017proximal} and its extension APO \cite{zhaoabsolute} ensure monotonic policy improvement. Another learning paradigm, imitation learning \cite{luo2023perpetual, he2024learning}, leverages large expert datasets to guide policy development. Despite their successes, these methods often suffer from challenges such as sample inefficiency, high computational cost, and limited generalization. 

Model-based methods, on the other hand, rely on leveraging system physics to construct accurate models. Techniques like iLQR \cite{li2004iterative} and NMPC \cite{grune2017nonlinear, liu2023modelbasedcontrolsparseneural} use these models to optimize control trajectories, while methods such as backstepping \cite{fossen1998nonlinear} are well-suited for systems with well-defined dynamics. Neural network-based approaches, such as \cite{nagabandi2017neuralnetworkdynamicsmodelbased}, extend these ideas by approximating nonlinear dynamics for control tasks. Moreover, model-based RL combines data-driven techniques with physical modeling to describe nonlinear system dynamics. For instance, \cite{gmm2021song} uses RL workflows to fit multiple local approximations of global dynamics, while \cite{nagabandi2017neuralnetworkdynamicsmodelbased} employs RL reward functions to learn dynamics from extensive random-shooting data. However, even these approaches face significant challenges when applied to high-dimensional systems, such as legged robots, where designing accurate models becomes increasingly complex and task-specific.

\textbf{Koopman Operator Theory.}
While model-based methods struggle with complex, high-dimensional systems and model-free methods lack interpretability and flexibility, linearizing nonlinear dynamics has recently emerged as a promising approach. Among these, Koopman Operator Theory has garnered attention for its ability to represent nonlinear dynamics in a higher-dimensional linear space. Current research on Koopman theory of control can be categorized into three main directions: 
\begin{enumerate}[leftmargin=15pt,itemsep=-5pt,  topsep=-10pt,label=(\roman*)]
\item  Koopman Operator as a Feature Encoder: The Koopman operator has been applied in some methods to predict future states based on the current state \cite{han2023utility} or vision inputs \cite{chen2024korol}. However, these approaches primarily serve as feature extractors and do not fully exploit the benefits of linearization. Additionally, designing an effective Koopman operator is often challenging due to the requirement for infinite-dimensional representations and the inefficiency of available data \cite{kaiser2020data}.
\item Integrating Koopman Operators with RL: Other methods combine the Koopman latent dynamic space with traditional RL policies, either hierarchically \cite{kim2024learning} or end-to-end \cite{lyu2023task}. While this integration aims to benefit from Koopman’s properties, it still inherits the challenges of learning-based methods, particularly in terms of interpretability and generalization
\item  Linear Control in Koopman Latent Space: A promising direction applies traditional linear control methods, such as LQR or MPC, to the Koopman-linearized model, emphasizing interpretability and flexibility. For example, \cite{shi2022deep} applies LQR to a deep-learned Koopman operator, while others use radial basis functions (RBF) \cite{korda2018linear} or derivative basis functions \cite{mamakoukas2021derivative} to design the operator. Despite these advances, current methods struggle to control high-dimensional systems, like legged robots due to severe model mismatch and domain shifts.
\end{enumerate}

Finally, research has begun addressing the errors that arise during the Koopman dynamics learning process. Works such as \cite{kim2024uncertainty} and \cite{korda2018convergence} evaluate the reconstruction and projection errors, providing valuable theoretical foundations for further improvement.

\textbf{Continual Learning.} Continual learning aims to incrementally update knowledge from streaming data, often utilized in learning prediction or dynamics models as seen in works like \cite{nagabandi2018deep,abuduweili2020robust}.  A prevalent approach in continual learning is memory replay \cite{rolnick2019experience}, where selected samples are stored and repeatedly replayed to train the model through methods such as random sampling \cite{rolnick2019experience} or importance sampling \cite{yin2023bioslam,abuduweili2023online}. From the memory replay perspective,  our approach adopts a random replay strategy to efficiently generate and reuse data.  Another relevant approach is the use of dynamic architecture-based methods, exemplified by Learning without Forgetting (LwF) \cite{li2017learning} and PackNet \cite{mallya2018packnet}, which modify the neural network architecture incrementally during training. 
In contrast to these methods, which typically add parameters in the final layers for task-specific learning, our method leverages a theory-guided strategy to expand the latent state dimension.
\section{Algorithm}
\Cref{alg:koopman} presents the proposed the Incremental Learning algorithm for Koopman Operators.

\begin{algorithm}
\caption{Incremental Koopman Algorithm }\label{alg:koopman}
\begin{algorithmic}

\STATE \textbf{Input:} MPC Controller $\pi_{mpc}$, Initial latent space dimension $n^{(0)}$, Initial dynamics $\mathcal{T}^{(0)}=(g^{(0)}_{n^{(0)}}, A^{(0)}_{n^{(0)}}, B^{(0)}_{n^{(0)}})$, Initial dataset $\mathcal{D}^{(0)}$, Reference repository $\mathcal{R}$, Dimension increment step size $\Delta n$  and Initial training epochs $J^{(0)}$

\STATE ~~~
\STATE \textbf{function:} TrainKoopman($n, \mathcal{D}, J$):
\begin{ALC@g}
\STATE $optimizaer \leftarrow$ Adam();
\STATE $scheduler \leftarrow$ CosineAnnealingLR();
\STATE $\mathcal{T} \doteq (g_n, A_n, B_n);$ \COMMENT{\textcolor{blue}{Initialization of Koopman dynamics}}
\FOR{$k=0,1,2,\dots,J-1$}
    \FOR{$\mathcal{B}$ in $\mathcal{D}$}
        \STATE $\mathcal{B}_z \leftarrow g(\mathcal{B});$ \COMMENT{\textcolor{blue}{Embed $x \in \mathcal{B}$ into latent vectors $z$}}
        \STATE $l_k \leftarrow \mathcal{L}_{koopman}(\mathcal{T}, \mathcal{B}, \mathcal{B}_z);$ \COMMENT{\textcolor{blue}{Refer to \Cref{eq: loss} for $\mathcal{L}_{koopman}$}}
        \STATE $optimizer$.zero\_grad();
        \STATE $l_k$.backward();
        \STATE $optimizer$.step();
    \ENDFOR
    \STATE $scheduler.step();$
\ENDFOR
\STATE $\textit{return~} \mathcal{T};$
\end{ALC@g}
\STATE \textbf{end function}
\STATE ~~~

\FOR{$j=0,1,2,\dots$}
    \STATE $n^{(j+1)} \leftarrow n^{(j)} + \Delta n;$ \COMMENT{\textcolor{blue}{Increase latent space dimension}}
    \STATE $\mathcal{D}^{(j+1)}_{incre} \leftarrow \pi_{mpc}(\mathcal{T}^{(j)}, \mathcal{R});$
    \STATE $\mathcal{D}^{(j+1)} \leftarrow \{\mathcal{D}^{(j)}, \mathcal{D}^{(j+1)}_{incre}\};$ \COMMENT{\textcolor{blue}{Dataset augmentation}}
    \STATE $J^{(j+1)} \leftarrow J^{(j)};$
    \STATE $\mathcal{T}^{(j+1)} \leftarrow$ TrainKoopman($n^{(j+1)}, \mathcal{D}^{(j+1)}, J^{(j+1)}$);
    \WHILE{TrainKoopman failed} 
        \STATE $J^{(j+1)} \leftarrow \frac{J^{(j+1)}}{2};$ \COMMENT{\textcolor{blue}{Adjust epochs when collapsing}}
        \STATE $\mathcal{T}^{(j+1)} \leftarrow$ TrainKoopman($n^{(j+1)}, \mathcal{D}^{(j+1)}, J^{(j+1)}$);
    \ENDWHILE
    \IF{TrackConverge$(\mathcal{T}^{(j+1)}, \mathcal{T}^{(j)})$}
        \STATE $Break;$ \COMMENT{\textcolor{blue}{Quit if tracking performance converge}}
    \ENDIF
\ENDFOR

\end{algorithmic}
\end{algorithm}
\section{Additional Experiments}
\label{sec: addi exp}

\subsection{Computational Cost Comparison}

\begin{wrapfigure}{t}{0.5\textwidth}
    \vspace{-10pt}
    \centering
    \includegraphics[width=0.48\textwidth]{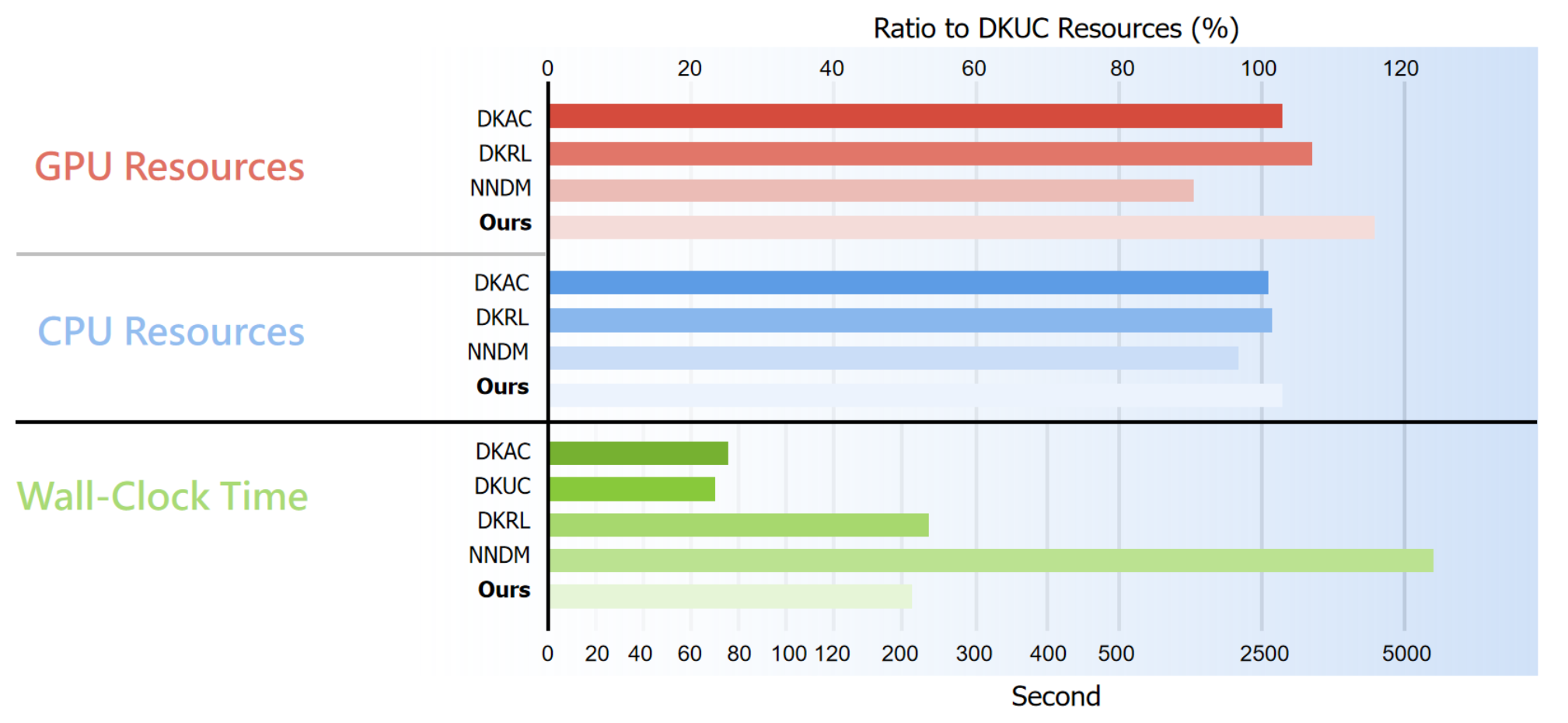}
    \caption{Computational cost(GPU occupancy, CPU occupancy and wall-clock time) comparison} 
    \label{fig: resources}
    \vspace{-10pt}
\end{wrapfigure}

We compare the resource usage of all algorithms in terms of GPU and CPU consumption, as well as wall-clock time, using DKUC as a benchmark, as presented in \Cref{fig: resources}. The horizontal axis in the upper area represents the percentage of resource usage relative to DKUC, while the horizontal axis in the lower area displays the running time (in seconds). The experiments are based on the averages from all seven test suites set for tracking experiments, as described in \Cref{sec: tracking}, and were run and averaged over three different seeds. Our algorithms utilize nearly the same GPU and CPU resources as the baseline methods while demonstrating a sharp lead in k-step prediction and tracking experiments. Furthermore, although wall-clock time increases when compared to DKUC and DKAC methods, it remains low enough to facilitate real-time inference for efficient control of real robots while NNDM method spends unacceptable reasoning time, demonstrating the effectiveness of the Koopman method and our algorithms, thereby addressing \textbf{Q5}.

\section{Proof of Koopman operator convergence}
\label{sec:sup_proof}
In this section, we discuss the theoretical properties of the proposed algorithm. The convergence of data-driven Koopman operator algorithms, particularly the Extended Dynamic Mode Decomposition (EDMD) \cite{Williams_Kevrekidis_Rowley_2015}, has been rigorously established in prior works \cite{korda2018convergence,korda2018linear}. Specifically, Korda and Mezić \cite{korda2018convergence} demonstrate that, under mild assumptions, as the sample size \( m \to \infty \) and the dimension of latent state (or also called number of observables) \( n \to \infty \), the estimated Koopman operator \( K \) converges to the true Koopman operator \( \mathcal{K} \). 
Our theoretical analysis builds upon the results presented in \cite{korda2018convergence,korda2018linear} but differs by providing explicit convergence rates under additional assumptions. In the following sections, we focus primarily on autonomous systems without control inputs. For control systems, the results can be extended by setting $s=[x;u]^\top$


\subsection{Preliminary}
\label{sec:preliminary}

    
    
    
    
\textbf{Koopman Operator Theory}.
For clarity, we adopt terminology slightly different from that used in the main content. Here $s^+$  denotes the next state of $s$.
Consider a dynamical system  $s^+=f(s)$, with $f: \mathcal{S} \rightarrow  \mathcal{S}$, where $ \mathcal{S}$ is a given separable topological space. Let $ \mathcal{H} $ represent a separable Hilbert space, equipped with an inner product \( \langle \cdot, \cdot \rangle \) and corresponding norm \( \| \cdot \| \).
The Koopman operator  $\mathcal{K}: \mathcal{H} \rightarrow  \mathcal{H}$ is defined as  $\mathcal{K} \phi = \phi \circ f$, where $\phi \in \mathcal{H}$ are embedding functions. 

\textbf{Extended Dynamic Mode Decomposition (EDMD)}.
EDMD is a data-driven method for approximating the Koopman operator $\mathcal{K}$ within a finite-dimensional subspace. Our methodology aligns with the EDMD approach.
Given a embedding function (observable) $\Phi(s)=[\phi_1(s), \phi_2(s), \dots, \phi_n(s)]^\top$, where $\phi_i : \mathcal{S} \rightarrow \mathbb{R} ~ \text{for}~ i \in [1,n]$. Given $m$ data pairs $\{(s_i, s_i^+)\}_{i=1}^m$,
EDMD seeks a finite-dimensional Koopman matrix $K_{n,m} \in \mathbb{R}^{n \times n}$ that best fits the data according to the following optimization problem:
\begin{align}
    \min_{K \in \mathbb{R}^{n \times n}} \sum_{i=1}^m \left\| \Phi(s_i^+) - K \Phi(s_i) \right\|_2^2 \label{eq:k_edmd}
\end{align}
Assuming that the Gram matrix $G_{n,m}$ is invertible, the closed-form solution  for the Koopman matrix is derived from:
\begin{align}
    K_{n,m} &=  A_{n,m} G_{n,m}^{-1} \label{eq:koopman_k}\\
    G_{n,m} &= \frac{1}{m} \sum_{i=1}^m \Phi(s_i) \Phi(s_i)^\top, ~ 
    A_{n,m} = \frac{1}{m} \sum_{i=1}^m \Phi(s_i) \Phi(s_i^+)^\top \label{eq:koopman_ga}
\end{align}
Denoting $\mathcal{H}_n$ the span of  $\phi_1(s), \phi_2(s), \dots, \phi_n(s)$. It is notable that any function $g \in \mathcal{H}_n $ can be represented as $g = c^\top \Phi$, where $c \in \mathbb{R}^n$. 

\textbf{EDMD as $L_2$ projection.} As shown by \cite{korda2018convergence}, EDMD approximates the finite-dimensional Koopman operator through an  $L_2$ projection of the true Koopman operator. We interpret these results as follows.  Consider an arbitrary nonnegative measure $\mu$ on the state space $\mathcal{S}$, and let $L_2(\mu)$ denote the Hilbert space of all square integrable functions with respect to the measure $\mu$. 
We define the $L_2(\mu)$ projection, denoted as $\mathcal{P}^{\mu}_n$ , of a  function $g$ onto $\mathcal{H}_n$:
\begin{align}
    \mathcal{P}^{\mu}_n g = \arg \min_{h \in \mathcal{H}_n} \| h - g \|_{L_2(\mu)} = \arg \min_{h \in \mathcal{H}_n} \int_{\mathcal{S}} (h - g)^2 \, d\mu = \Phi^\top \arg \min_{c \in \mathbb{R}^n} \int_{\mathcal{S}} (c^\top \Phi - g)^2 \, d\mu.
\end{align}
We now state a key property of $K_{n,m}$ as follows.

\textit{
\textbf{Theorem A.1 \cite{korda2018convergence}}. Let $ \hat{\mu}_m $ denote the empirical measure associated to the points $ s_1, \dots, s_m $, i.e., $ \hat{\mu}_m = \frac{1}{m} \sum_{i=1}^m \delta_{s_i} $, where $ \delta_{s_i} $ denotes the Dirac measure at $ s_i $. Then for any $ g \in \mathcal{H}_N $,
\begin{align}
    {K}_{n,m} g = \mathcal{P}^{\hat{\mu}_m}_n \mathcal{K} g = \arg \min_{h \in \mathcal{H}_n} \| h - \mathcal{K} g \|_{L_2(\hat{\mu}_m)}, ~ \text{~i.e.~} {K}_{n,m} = \mathcal{P}^{\hat{\mu}_m}_n \mathcal{K} |_{\mathcal{H}_n}, \label{eq:thoerom_koopman_proj}
\end{align}
where  $ \mathcal{K} |_{\mathcal{H}_n} : \mathcal{H}_n \to \mathcal{H}  $ is the restriction of the Koopman operator to the subspace  $ \mathcal{H}_n  $.
}

The theorem states that the estimated Koopman operator $\mathcal{K}_{n,m} $ is the $  L_2 $  projection of the true Koopman operator \( \mathcal{K} \) on the span of $ \phi_1, \dots, \phi_n$  with respect to the empirical measure on the samples $ s_1, \dots, s_m $. 

Consequently, the total error in approximating the true Koopman operator $ \mathcal{K} $ with its learned finite-dimensional estimate $ K_{n,m} $ can be decomposed into two components: projection error and sampling error.
\textit{Sampling error}: The sampling error,  denoted as $\epsilon_{\text{samp}}(n, m) = \| K_{n,m} - K_n\|$, originates from estimating the finite-dimensional Koopman operator $ K_n $ using a finite number of data samples $ m $.
\textit{Projection error}: The projection error, denoted as $\epsilon_{\text{proj}}(n) = \| K_n \mathcal{P}_n^\mu - \mathcal{K} \|$, arises from the approximation of the infinite-dimensional Koopman operator $\mathcal{K}$ within a finite-dimensional subspace. 
This conceptual framework forms the basis for the overall proof scheme presented in prior work by Korda and Mezić \cite{korda2018convergence}. In contrast, we extend these results by providing explicit convergence rates under certain additional assumptions.

\subsection{Convergence of Sampling error}
We begin by establishing the assumptions necessary for our analysis.

\noindent \textit{
\textbf{Assumption 1 (i.i.d samples)}. Data samples $s_1, \cdots, s_m$ are drawn independently from the distribution $\mu$. }

\noindent \textit{
\textbf{Assumption 2 (Bounded latent state)}. The norm of the latent state vector is bounded. Specifically, there exists a constant $ B > 0 $ such that  $\| \Phi(s) \| \le B$ for all $s \in \mathcal{S}$. }

\noindent \textit{
\textbf{Assumption 3 (Invertable Gram matrix)}. The Gram matrix $G_{n,m}$ defined in $ \cref{eq:koopman_ga}$ is invertible, and its smallest eigenvalue is bounded below by $\gamma > 0$, such as $\lambda_{\min}(G_{n,m}) \geq \gamma$.}

Assumption 1 ensures the independence of samples, which can be relaxed if the dynamical system  $f$ is ergodic and the samples  $s_1, \cdots, s_m$ are taken along a trajectory of the system starting from some initial condition  $s_0 \in \mathcal{S}$.  
Under Assumption 3, the learned Koopman operator $K_{n,m}$ can be can be computed using \cref{eq:koopman_k}. We consider the sampling error between the learned Koopman operator $K_{n,m}$ with finite data size $m$, and the Koopman operator $K_n$ obtained as the $L_2$ projection of the true Koopman operator onto the subspace $\mathcal{H}_n$) with infinite data:  
\begin{align}
   & \epsilon_{\text{samp}}(n, m)  = \| K_{n,m} - K_{n} \| = \| G_{n,m}^{-1} A_{n,m} -   G_{n}^{-1} A_{n} \| 
\end{align} 
Here, $G_{n,m}$ and $A_{n,m}$ are computed using finite samples via \cref{eq:koopman_ga}. While  $G_n$ and $A_n$, represent their expected values, equivalent to computing \cref{eq:koopman_ga} with an infinite number of samples. Under Assumption 1, by the Law of Large Numbers,  $G_{n,m}$ and $A_{n,m}$ converge to the expected value $G_{n}$ and $A_{n}$ respectively as $m \rightarrow \infty$. Then we have:
\begin{align}
    & G_n  = \mathbb{E}[G_{n,m}] =  \mathbb{E}\left[ \frac{1}{m} \sum_{i=1}^m \Phi(s_i) \Phi(s_i)^\top \right], ~~ A_n = \mathbb{E}[A_{n,m}] = \mathbb{E} \left[ \frac{1}{m} \sum_{i=1}^m \Phi(s_i) \Phi(s_{i}^+)^\top \right]
\end{align}
We proceed to decompose the sampling error:
\begin{align}
\epsilon_{\text{samp}}(n, m) & = \|  A_{n,m}G_{n,m}^{-1} -   A_{n} G_{n}^{-1} \| 
= \| A_{n,m} G_{n,m}^{-1} - A_{n,m} G_{n}^{-1} + A_{n,m}G_{n}^{-1}  - A_{n} G_{n}^{-1} \| \nonumber \\
&\leq \| A_{n,m} \| \cdot \| G_{n,m}^{-1} - G_{n}^{-1} \| +  \| A_{n,m} - A_{n}  \| \cdot \|G_{n}^{-1} \| . \label{eq:theory_est_err1}
\end{align}
To bound each term in  \cref{eq:theory_est_err1}. , we utilize properties of matrix norms and inequalities. First, we consider the difference of the inverses of $G_{n,m}$ and $G_n$. Using the identity $ G_{n,m}^{-1} - G_{n}^{-1} = \ G_{n}^{-1} ( G_{n} -  G_{n,m})  G_{n,m}^{-1}$, we have:
\begin{align}
\|  G_{n,m}^{-1} - G_{n}^{-1} \| & \leq \| G_{n}^{-1} \| \| G_{n} - G_{n,m} \| \|G_{n,m}^{-1}  \|. \label{eq:theory_matI}
\end{align}
Based on the assumption 2, the spectral norm of $A_{m,n}$ can be bounded:
\begin{align}
    & \| A_{n,m} \|  \le  \frac{1}{m} \sum_{t=1}^m \|\Phi(s_i)\| \|\Phi(s_i^+)\| \le B^2 \label{eq:theory_anorm}
\end{align}
Similarly, under Assumption 3, the norms of inverse gram matrices are bounded
\begin{align}
 \| G_{n,m}^{-1} \| = \frac{1}{\lambda_{\min}(G_{n,m})} \le \frac{1}{\gamma},  ~ \| G_{n}^{-1} \| \le \frac{1}{\gamma}\label{eq:theory_inverse_norm}
\end{align}
Substituting\cref{eq:theory_matI,eq:theory_anorm,eq:theory_inverse_norm} back to \cref{eq:theory_est_err1}, we obtain:
\begin{align}
\epsilon_{\text{samp}}(n, m) & \leq \frac{B^2 }{\gamma^2} \| G_{n,m} -G_{n} \| +  \frac{1}{\gamma} \|A_{n,m} - A_n   \| \label{eq:thoery_est_err2}
\end{align}

To bound $\|G_{n,m} -G_{n} \|$ and $\| A_{n,m} - A_n   \|$,we utilize the Matrix Bernstein Inequality \cite{tropp2012user}, which provides concentration bounds for sums of independent random matrices.

\textit{Matrix Bernstein Inequality.
Let \( \{ Y_i \}_{i=1}^m \) be independent, mean-zero random matrices with dimensions \( d_1 \times d_2 \). Assume that each matrix satisfies \( \| Y_i \| \leq L \) almost surely. Define the variance parameter}
\begin{align}
    \sigma^2 = \left\| \sum_{i=1}^m \mathbb{E}[ Y_i Y_i^\top ] \right\|.
\end{align}
\textit{Then, for all \( \epsilon \geq 0 \),}
\begin{align}
\mathbb{P}\left( \left\| \sum_{i=1}^m Y_i \right\| \geq \epsilon \right) \leq (d_1 + d_2) \exp\left( \frac{ -\epsilon^2 / 2 }{ \sigma^2 + L \epsilon / 3 } \right).
\end{align}
We apply the Matrix Bernstein Inequality to bound $ \| A_{n,m} - A_n \|$. Define $Y_i$ as:
\begin{align}
& Y_i = \frac{1}{m} \left(  \Phi(s_i) \Phi(s_{i}^+)^\top -  \mathbb{E}[\Phi(s_i) \Phi(s_{i}^+)]^\top   \right).
\end{align}
Then $A_{n,m} - A_n  = \sum_{i=1}^m Y_i$, and $E[Y_i] = 0$.
Under Assumption 2, the norm of  $Y_i$ is bounded:
\begin{align}
    \| Y_i \| \leq \frac{2}{m} \cdot \| \Phi(s_i) \| \cdot \| \Phi(s_{i}^+) \| \leq \frac{2 B^2}{m}
\end{align}
he variance parameter $\sigma^2$ is bounded as follows:
\begin{align}
    \sigma^2 &= \left\| \sum_{i=1}^m \mathbb{E}[ Y_i Y_i^\top ] \right\| \le \sum_{i=1}^m \| \mathbb{E} \left[ Y_i Y_i^\top \right]   \|  \le \sum_{i=1}^m \left(  \frac{2B^2}{m} \right)^2 = \frac{4B^4}{m}
\end{align}
Applying the Matrix Bernstein Inequality with $d_1=d_2=n, L =\frac{2B^2}{m}, \sigma^2= \frac{4B^4}{m}$, we obtain:
\begin{align}
\mathbb{P}\left( \| A_{n,m} - A_n \| \geq \epsilon \right) &= \mathbb{P}\left( \left\| \sum_{i=1}^m Y_i \right\| \geq \epsilon \right) 
  \leq 2n \exp\left( \frac{ -\epsilon^2 m / 2 }{ 4 B^4 + \frac{2 B^2 \epsilon}{3} } \right).
\end{align}
Assuming  $\epsilon < B^2$, and for sufficiently large $m$,  the term $ 4 B^4$ dominates $\frac{2 B^2 \epsilon}{3} $, allowing us to simplify the bound::
\begin{align}
\mathbb{P}\left( \|  A_{n,m} - A_n \| \geq \epsilon \right) \leq 2n \exp\left( - \frac{m \epsilon^2}{8 B^4} \right).
\end{align}
This bound holds in probability. To ensure that the probability is at most $ \delta $, we set:
\begin{align}
2n \exp\left( - \frac{m \epsilon^2}{8 B^4} \right) \leq \delta.
\end{align}
Solving for $ \epsilon $, we find:
\begin{align}
\epsilon \geq \sqrt{ \frac{8 B^4 \ln(2n / \delta)}{m} }.
\end{align}
Thus, with probability at least $ 1 - \delta $, the following inequality holds,
\begin{align}
\|  A_{n,m} - A_n \| \leq \sqrt{ \frac{8 B^4 \ln(2n / \delta)}{m} }. \label{eq:theory_a_ea_bound}
\end{align}
A similar procedure applies to bound $\|  G_{n,m} - G_n \|$.  Using the Matrix Bernstein Inequality with probability at least \( 1 - \delta \),
\begin{align}
\| G_{n,m} -G_{n} \| \leq \sqrt{ \frac{8 B^4 \ln(2n / \delta)}{m} }. \label{eq:theory_g_eg_bound}
\end{align}
Substituting \cref{eq:theory_a_ea_bound,eq:theory_g_eg_bound} into \cref{eq:thoery_est_err2}, we obtain, with probability at least$ (1 - \delta)^2 $, 
\begin{align}
    \epsilon_{\text{samp}}(n, m) & \leq \frac{B^2 }{\gamma^2} \sqrt{ \frac{8  B^4 \ln(2n / \delta)}{m} } +  \frac{1}{\gamma} \sqrt{ \frac{8 B^4 \ln(2n / \delta)}{m} } =   \ \frac{2 B^2  \sqrt{2 \ln(2n / \delta)}}{\gamma \sqrt{m}} \left( \frac{B^2}{\gamma} + 1\right)  \label{eq:thoery_est_err3}
\end{align}

For simplicity, we mainly consider the convergence rate related to $m,n$, and then with a high probability,
\begin{align}
  \epsilon_{\text{samp}}(n, m) & \leq \mathcal{O}( \sqrt{ \frac{\ln(n)}{m} })  \label{eq:converge_sample_err_rate}
\end{align}
For larger sample sizes $m$, such that $m=\mathcal{O}\left(n \ln (n)\right)$, with high probability we have:
\begin{align}
    \epsilon_{\text{samp}}(n, m) |_{m=\mathcal{O}\left(n \ln (n)\right)} & \leq \mathcal{O}(\frac{1}{\sqrt{n}}) \label{eq:converge_kmn}
\end{align}
This result indicates that, under the given assumptions, the sampling error decreases inversely with the square root of the number of observables $n$, when the number of samples $m$ grows proportionally to $n \ln n$.

\subsection{Convergence of Projection Error}
In this section, we establish the convergence of the projection error $\epsilon_{\text{proj}}(n)$  as the dimension $n$ of the subspace $\mathcal{H}_n$ increases. We begin by introducing additional assumptions necessary for our analysis.

\noindent \textit{
\textbf{Assumption 4 (Orthogonal basis)}. The embedding functions $\phi_1, \cdots, \phi_n$ are orthogonal (independent). i.e. $ \langle \phi_i, \phi_j \rangle = \delta_{ij}, ~ \text{for all } i, j = 1, \dots, n, $ where $\delta_{ij}$ is the Kronecker delta.}

\noindent \textit{
\textbf{Assumption 5 (Bounded Koopman Operator)}. The Koopman operator $\mathcal{K}: \mathcal{H} \rightarrow \mathcal{H}$ is bounded, i.e. $ \| \mathcal{K} \| \le M < \infty $.}


Given a function $\psi \in \mathcal{H}$,  we define the projection error as
\begin{align}
\epsilon_{\text{proj}}(n) =\| K_n \mathcal{P}_n^\mu \psi - \mathcal{K} \psi \| =  \int_{\mathcal{S}} \|K_n \mathcal{P}_n^\mu \psi - \mathcal{K} \psi \| d \mu 
\end{align}
where $K_n$ is the finite-dimensional approximation of the Koopman operator defined on $\mathcal{H}_n$, and $\mathcal{P}_n^\mu$ is the orthogonal projection onto $\mathcal{H}_n$.

We decompose $\psi$ into its projection onto $\mathcal{H}_n$ and its orthogonal complement: $\psi = \mathcal{P}_n^\mu \psi + (I -\mathcal{P}_n^\mu ) \psi $, where $I$ is the identity operator. Then we have:
\begin{align}
\epsilon_{\text{proj}}(n) &= \| K_n \mathcal{P}_n^\mu \psi - \mathcal{K} \psi \| \\ 
&=\| K_n \mathcal{P}_n^\mu  \mathcal{P}_n^\mu \psi - \mathcal{K}  \mathcal{P}_n^\mu \psi +  K_n \mathcal{P}_n^\mu  (I -\mathcal{P}_n^\mu) \psi -  \mathcal{K}  (I -\mathcal{P}_n^\mu) \psi \|  \label{eq:decopose_psi}
\end{align}
According to the definition of  the orthogonal projection $\mathcal{P}_n^\mu$, 
we have $ \mathcal{P}_n^\mu  \mathcal{P}_n^\mu \psi = \mathcal{P}_n^\mu \psi, ~ \mathcal{P}_n^\mu \cdot (I - \mathcal{P}_n^\mu) \psi = 0$. Recall from Theorem A.1 that $K_n = \mathcal{P}_n^\mu \mathcal{K}$. Therefore, we can rewrite \cref{eq:decopose_psi} as :
\begin{align}
\epsilon_{\text{proj}}(n) &= \| K_n \mathcal{P}_n^\mu  \psi - \mathcal{K}  \mathcal{P}_n^\mu \psi  -  \mathcal{K}  (I -\mathcal{P}_n^\mu) \psi \| \\
& = \| (\mathcal{P}_n^\mu - I ) \mathcal{K} \mathcal{P}_n^\mu  \psi  +  \mathcal{K}  (\mathcal{P}_n^\mu - I) \psi \| \\
& \leq \| (\mathcal{P}_n^\mu - I) \| \| \mathcal{K} \mathcal{P}_n^\mu \psi \| + \| \mathcal{K} \| \| (\mathcal{P}_n^\mu - I) \| \| \psi \|  \label{eq:decopose_psi_2}
\end{align}
Under Assumption 5, $\|\mathcal{K} \| \le M$. Under assumption 2, $\|\phi_i \| \le B$. Let $\psi = \sum_{i=1}^\infty c_i \phi_i$. By Parseval’sidentity $\sum_{i=1}^\infty |c_i|^2 = 1$. Then  $\|\psi \| \le B$. Therefore, we can rewrite \cref{eq:decopose_psi_2} as :
\begin{align}
   \epsilon_{\text{proj}}(n) \leq  2 M B \| (\mathcal{P}_n^\mu - I) \| \label{eq:simplified_error_bound}
\end{align}
From Equation~\eqref{eq:simplified_error_bound}, we observe that the projection error depends on the norm of the residual $  (\mathcal{P}_n^\mu - I) \psi$. As $n \rightarrow \infty$,  the subspace $\mathcal{H}_n$ becomes dense in $\mathcal{H}$, and $\mathcal{P}_n^\mu$ converges strongly to the identity operator $I$ \cite{korda2018convergence}. Therefore,
\begin{align}
    \lim_{n \rightarrow \infty} \epsilon_{\text{proj}}(n) =0, ~  \lim_{n \rightarrow \infty} K_n \mathcal{P}_n^\mu = \mathcal{K} \label{eq:converg_kn}
\end{align}

Recall that $K_n$ is an operator on the finite-dimensional subspace $\mathcal{H}_n$ spanned by the embedding functions  $\phi_1, \cdots, \phi_n$. And $\mathcal{K}$ is an operator on $\mathcal{H}$. The convergence of $K_n$ to $\mathcal{K} $ implies convergence of their eigenvalues and eigenfunctions under certain conditions. Prior work \cite{korda2018convergence} shows the convergence of eigenvalue as the following theorem.  

\textit{
\textbf{Theorem A.2 \cite{korda2018convergence}}. 
If assumption 4 and assumption 5 holdes, and $\lambda_n$ is a sequance of eigenvalue of $K_n$ with associated normalized eigenfunctions $\phi_n \in \mathcal{H}_n$, then there exists a subsequence $(\lambda_{n_i}, \phi_{n_i})$ such that
\begin{align}
\lim_{i \rightarrow \infty} \lambda_{n_i} = \lambda^\star, ~  \phi_{n_i} \xrightarrow{\text{weak convergence}} \phi^\star
\end{align}
where $\lambda^\star \in \mathbb{C}$ and $\phi^\star \in \mathcal{H}$ are eigenvalue and eigenfunction of Koopman operator, such that $\mathcal{K} \phi^\star = \lambda^\star \phi^\star$.
}

This theorem demonstrates the convergence of eigenvalues and eigenfunctions of $K_n$ to those of $\mathcal{K} $ as $n \rightarrow \infty$. 
Let the sorted (decreasing order) eigenvalues of the Koopman operator $\mathcal{K}$ are $\lambda_1^\ast, \lambda_2^\ast, \cdots$, and corresponding eigenfunctions are $\phi_1^\ast, \phi_2^\ast, \cdots$. The sorted eigenvalues of the $K_n$ are $\lambda_1, \lambda_2, \cdots, \lambda_n$, and corresponding eigenfunctions are $\phi_1, \phi_2, \cdots, \phi_n$.
To provide an explicit convergence rate for the projection error, we introduce further assumptions.

\noindent \textit{
\textbf{Assumption 6 (Spectral Decay)}. The eigenvalues of the Koopman operator $\mathcal{K}$ decay sufficiently fast, for instance, $ | \lambda_i | \leq \frac{C}{i} $ for some constant $ C > 0 $.}

\noindent \textit{
\textbf{Assumption 7 (Accurate Learning of Dominant Modes)}.  The embedding functions $\Phi(\cdot) = [\phi_1, \cdots,\phi_n]^\top$ correspond to the first $n$ eigenfunctions of $\mathcal{K}$ associated with the largest eigenvalues in magnitude. That is, $\lambda_i = \lambda_i^\ast$ or   $\langle \lambda_i \phi_i, f \rangle = \langle \lambda_i^\ast \phi_i^\ast, \mathcal{P}_n^\mu f \rangle $ for $f \in \mathcal{H}$.  
}

Assumption 6 holds for systems where the Koopman operator has rapidly decaying spectral components.
Under assumption 7, we can quantify the convergence rate of $\epsilon_{\text{proj}}(n)$.  
Let $\psi \in \mathcal{H}$, be expressed in terms of the eigenfunctions of $\mathcal{K}$: $\psi = \sum_{i=1}^\infty c_i \phi_i^\star \in \mathcal{H}$ with $\sum_{i=1}^\infty |c_i|^2 = 1$.  The projection error is given by
\begin{align}
\epsilon_{\text{proj}}(n) &= \| K_n \mathcal{P}_n^\mu \psi - \mathcal{K} \psi \| =\|    \sum_{i=1}^\infty c_i K_n \mathcal{P}_n^\mu \phi_i^\star  -  \sum_{i=1}^\infty c_i \mathcal{K} \phi_i^\star \| \\
&= \|   \sum_{i=1}^n c_i \lambda_i \phi_i^\star  -  \sum_{i=1}^\infty c_i \lambda_i^\star \phi_i^\star \|  = \| \sum_{i=1}^n c_i (\lambda_i - \lambda_i^\star ) \phi_i^\star +  \sum_{i=n+1}^\infty c_i \lambda_i^\star \phi_i^\star \| \\
& = \|   \sum_{i=n+1}^\infty c_i \lambda_i^\star \phi_i^\star \|  \le B \cdot  \left( \sum_{i=n+1}^\infty | \lambda_i^\star |^2 \right)^{1/2} 
\end{align}
Considering assumption 6,  we can estimate the projection error
\begin{align}
\epsilon_{\text{proj}}(n) &\leq B  \left( \sum_{i=n+1}^\infty \left( \frac{C}{i} \right)^2 \right)^{1/2}   
= BC  \left( \sum_{i=n+1}^\infty \frac{1}{i^2}  \right)^{1/2}  \\
& \le   BC  \left( \sum_{i=n+1}^\infty \left( \frac{1}{i-1} - \frac{1}{i}  \right)\right)^{1/2}  = \frac{BC}{\sqrt{n}}. 
\end{align}
Therefore, we have the convergence rate of projection error:
\begin{align}
\epsilon_{\text{proj}}(n) \leq \mathcal{O}\left( \frac{1}{\sqrt{n}} \right) \label{eq:proj_err_converg}
\end{align}
Thus, under the given assumptions, the projection error decreases inversely with the square root of $n$. 


\subsection{Convergence of Koopman Operator Under Incremental Strategy}
\label{sec:ap_theory_final}
By combining the convergence results of the sampling error \cref{eq:converge_kmn} and the projection error\cref{eq:converg_kn}, we conclude that the estimated Koopman operator $K_{m,n}$ converges to the true Koopman operator $\mathcal{K}$. This convergence has also been established in previous work by Korda and Mezić \cite{korda2018convergence}.

\textit{
\textbf{Theorem A.3} Under the assumptions of 1) data samples $s_1, \cdot, s_m$ are i.i.d distributed; 2) the latent state is bounded, $\|\phi(s)\| < \infty$; 3) the embedding functions $\phi_1, \cdots, \phi_n$ are orthogonal (independent). 
The learned Koopman operator $K_{m,n}$ converges to true Koopman Operator:
\begin{align}
    \lim_{m=\Omega(n ln(n)), n \rightarrow \infty} K_{m,n} \rightarrow \mathcal{K}
\end{align}. 
}

Unlike prior work, we provide explicit convergence rates for both the sampling error and the projection error \cref{eq:converge_kmn,eq:proj_err_converg}, leading to an overall error bound for $K_{m,n}$.

\textit{
\textbf{Theorem A.4} Under the assumptions of 1) data samples $s_1, \cdot, s_m$ are i.i.d distributed; 2) the latent state is bounded, $\|\phi(s)\| < \infty$; 3) the embedding functions $\phi_1, \cdots, \phi_n$ are orthogonal. \\
(a) \textbf{Sampling Error Rate:} The sampling error decreases with the number of samples $m$ as:
\begin{align}
 \epsilon_{\text{samp}}(n, m) & \leq \mathcal{O}( \sqrt{ \frac{\ln(n)}{m} })
\end{align} 
(b) \textbf{Projection Error Rate:} Assuming additional conditions: 4) The eigenvalues of the Koopman operator $\mathcal{K}$ decay sufficiently fast, for instance, $ | \lambda_i | \leq \frac{C}{i} $ for some constant $ C > 0 $. 5) The embedding functions $\Phi(\cdot) = [\phi_1, \cdots,\phi_n]^\top$ correspond to the first $n$ eigenfunctions of $\mathcal{K}$ associated with the largest eigenvalues in magnitude.Then, the projection error decreases with $n$:
\begin{align}
\epsilon_{\text{proj}}(n) \leq \mathcal{O}\left( \frac{1}{\sqrt{n}} \right)
\end{align} 
(c) \textbf{Overall Error Bound:} The total approximation error between $K_{m,n}$ and true koopman operator $\mathcal{K}$ satisfies:
\begin{align}
\text{error} \leq \mathcal{O}( \sqrt{ \frac{\ln(n)}{m} })+  \mathcal{O}\left( \frac{1}{\sqrt{n}} \right)
\end{align} 
}

\section{Expeiment Details}
\label{sec: appendix exp}

\subsection{Task Settings}
\label{sec: task settings}

\begin{figure}[h]
    \centering
    \subfigure[ANYmal-D]{
        \centering
        \includegraphics[width=0.16\textwidth]{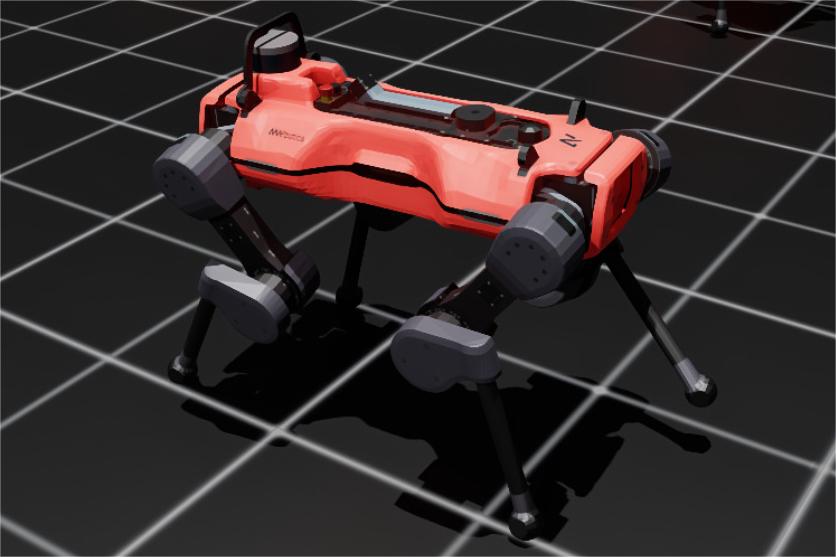}
        \label{fig:anymal-d}
    }
    \subfigure[Unitree-A1]{
        \centering
        \includegraphics[width=0.16\textwidth]{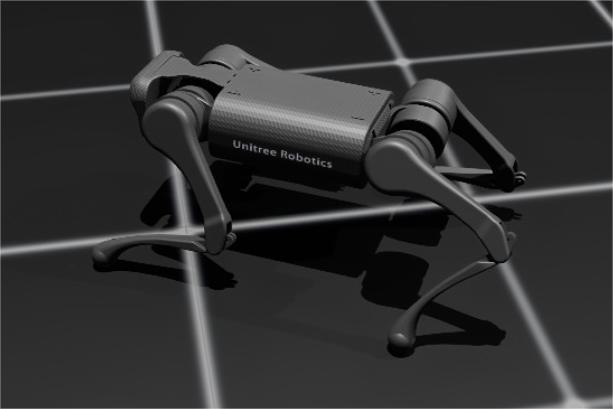}
        \label{fig:a1}
    }
    \subfigure[Unitree-Go2]{
        \centering
        \includegraphics[width=0.16\textwidth]{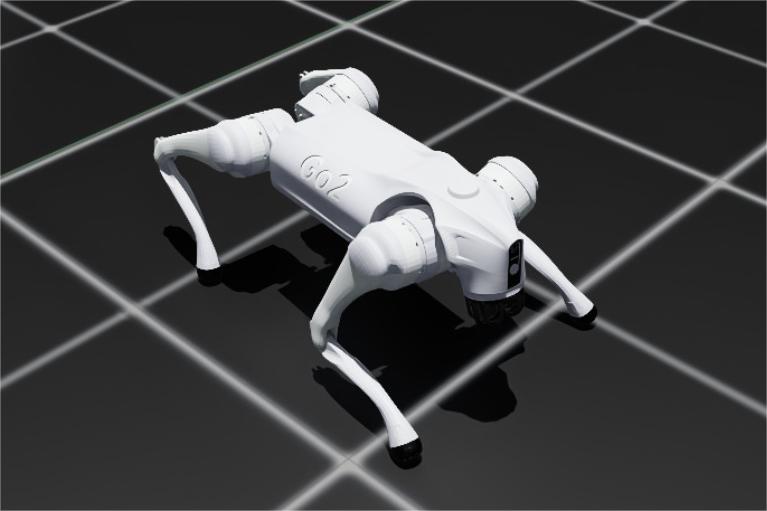}
        \label{fig:go2}
    }
    \subfigure[Unitree-H1]{
        \centering
        \includegraphics[width=0.16\textwidth]{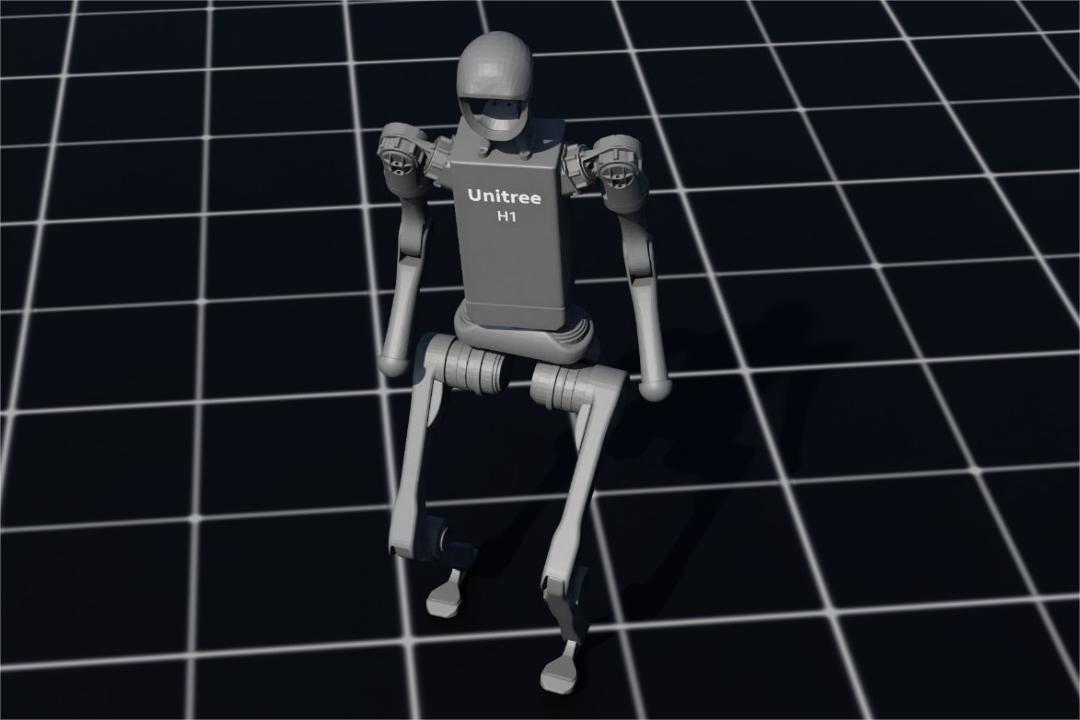}
        \label{fig:h1}
    }
    \subfigure[Unitree-G1]{
        \centering
        \includegraphics[width=0.16\textwidth]{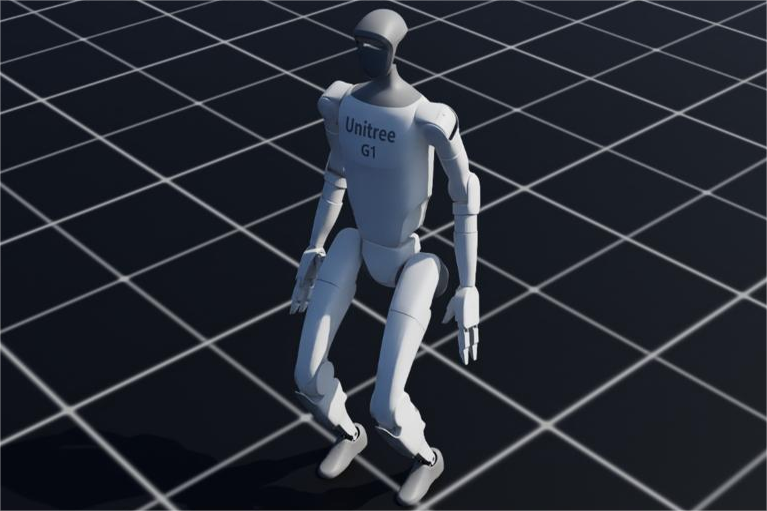}
        \label{fig:g1}
    }
    \subfigure[Flat]{
        \centering
        \vspace{10pt}
        \includegraphics[width=0.16\textwidth]{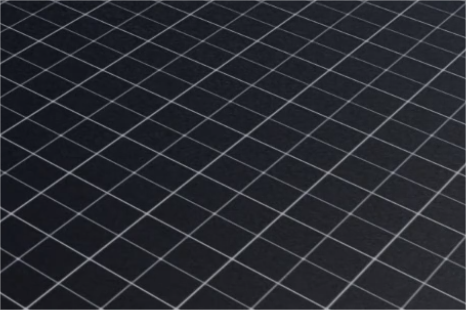}
        \label{fig:flat}
    }
    \subfigure[Rough]{
        \centering
        \vspace{10pt}
        \includegraphics[width=0.16\textwidth]{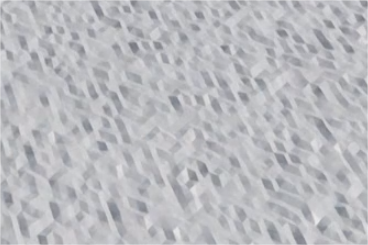}
        \label{fig:rough}
    }
    \caption{Legged robots and terrain in our test suites.}
    \label{fig: robots}
    \vspace{-10pt}
\end{figure}

We summarize our test suites in \Cref{tab: test suites}, and present the tracking failure threshold $\epsilon_{fail}$ in \Cref{tab: hyperparameter for task suites}. It is worth noting that robots performing different tasks have distinct physical structures and movement patterns, necessitating different $\epsilon_{fail}$ values. These hyperparameters are selected when the robot is on the verge of falling and entering an unrecoverable state.

\begin{table}[h]
\vskip 0.15in
\caption{The settings of test suites environment for our experiments}
\begin{center}
\begin{tabular}{cc|cc|cc|cc}
\toprule
 \multicolumn{2}{c|}{\textbf{Task Settings}} & \multicolumn{2}{c|}{Producer} &\multicolumn{2}{c|}{Terrain} &\multicolumn{2}{c}{Robot Type}\\
\cline{3-8}\\[-1.02em]
& & Unitree & ANYmal & Flat & Rough & Quadruped & Humanoid \\
\cline{1-8}\\[-0.99em]
\multicolumn{1}{c|}{} & Anymal-D ($\mathbb{R}^{12}$)   &  & \checkmark & \checkmark &&\checkmark& \\
\multicolumn{1}{c|}{IsaacLab} & A1 ($\mathbb{R}^{12}$)   &  \checkmark &  &\checkmark & &\checkmark&  \\
\multicolumn{1}{c|}{} & Go2 ($\mathbb{R}^{12}$)  & \checkmark &  &\checkmark &\checkmark&\checkmark& \\
\multicolumn{1}{c|}{Robot} & H1 ($\mathbb{R}^{19}$)    & \checkmark &  &\checkmark & && \checkmark\\
\multicolumn{1}{c|}{} & G1 ($\mathbb{R}^{23}$)      & \checkmark &  &\checkmark & \checkmark&&\checkmark \\
\bottomrule
\end{tabular}
\label{tab: test suites}
\end{center}
\end{table}

\begin{table}[h]
\vskip 0.15in
\caption{The settings of hyperparameter $\epsilon_{fail}$ for test suites}
\begin{center}
\begin{tabular}{cc|ccccc}
\toprule
 \multicolumn{2}{c|}{\textbf{$\epsilon_{fail}$ Settings}} & Anymal-D & A1 & Go2 & H1 & G1 \\
\cline{1-7}\\[-0.99em]
\multicolumn{1}{c|}{Terrain} & Flat & 0.18 & 0.16 & 0.16 & 0.15 & 0.10 \\
\multicolumn{1}{c|}{Type} & Rough & - & - & 0.12 & - & 0.08 \\
\bottomrule
\end{tabular}
\label{tab: hyperparameter for task suites}
\end{center}
\end{table}

\subsection{Dataset Settings}
\label{sec: dataset settings}
All five legged robots are trained on IsaacLab \cite{mittal2023orbit}. The initial dataset $\mathcal{D}^{(0)}$ consists of 6e4 trajectories of length $l_{\mathcal{D}^{(0)}}$, and the reference repository $\mathcal{R}$, containing 3e3 demonstrations of length $l_{\mathcal{R}}$ showcasing diverse and rich walking styles.

To achieve a more uniform data distribution under specific gait and contact modes, we apply random initialization techniques for constructing $\mathcal{D}^{(0)}$ and $\mathcal{R}$. First, we randomize the initial movement commands, including heading direction, x-axis linear velocity, and y-axis linear velocity in world coordinates, where the x-axis aligns with the heading direction and the y-axis is perpendicular to it. After collecting 6e4 trajectories of length $l_{init}$ (where $l_{init}$ > $l_{\mathcal{D}^{(0)}}$), we randomly extract continuous segments of length $l_{\mathcal{D}^{(0)}}$ to enrich the robot's range of motion and postures.  All detailed hyperparameters are provided in \Cref{tab: algo settings}.

In the following lifting phase, incremental dataset $\mathcal{D}_{incre}$, containing 3e4 trajectories of length $l_{\mathcal{D}^{(0)}}$, will be added to the training dataset.

The test dataset for evaluating k-step prediction error consists of 30,000 "success" data points and 30,000 "fail" data points. The success data is collected using the initial data collector with a stable walking gait, while the fail data is generated by all five algorithms tested in our experimental setup. Specifically, each algorithm produces 6,000 trajectories using the incremental data collector. Iterative algorithms, such as the Incremental Koopman Algorithm and DKRL, generate testing data based on the initial model. Notably, the test dataset is constructed independently of the lifting phase, ensuring that the test data is not used for re-training the dynamics.

\subsection{Algorithm Settings}
\label{sec: algo settings}

\begin{sidewaystable}
\vskip 0.15in
\caption{Important hyperparameters of different algorithms in our experiments}
\begin{center}
\resizebox{\textwidth}{!}{%
\begin{tabular}{lr|ccccc}
\toprule
\textbf{Policy Parameter} & & \textbf{Ours} & DKRL & DKUC & DKAC & NNDM\\
\hline\\[-1.0em]
Original space dimension & $n'$ & (35, 35, 35, 35, 45, 53, 53) & (35, 35, 35, 35, 45, 53, 53) & (35, 35, 35, 35, 45, 53, 53) & (35, 35, 35, 35, 45, 53, 53) & (35, 35, 35, 35, 45, 53, 53) \\
Control input dimension & $m'$ & (12,12,12,12,19,23,23) & (12,12,12,12,19,23,23) & (12,12,12,12,19,23,23) & (12,12,12,12,19,23,23) & (12,12,12,12,19,23,23) \\
Initial latent space dimension & $n^{(0)}$ &(512, 384, 384, 535, 214, 222, 453) &(512, 384, 384, 535, 214, 222, 453) &(512, 384, 384, 535, 214, 222, 453) &(512, 384, 384, 535, 214, 222, 453) &(512, 384, 384, 535, 214, 222, 453) \\
Fixed step size & $\Delta n$ &100 &100 &100 &100 &100 \\
Inference horizon & H &(16,16,16,16,24,16,16) &(16,16,16,16,24,16,16) &(16,16,16,16,24,16,16) &(16,16,16,16,24,16,16) &(16,16,16,16,24,16,16) \\

Initial training epochs & $J^{(0)}$ & 100 & 100 & 100 & 100 & 100 \\
Initial learning rate & &1e-3 &1e-3 &1e-3 &1e-3 &1e-3 \\
Network optimizer & &Adam &Adam &Adam &Adam &Adam \\
Network scheduler & &CosineAnnealinLR &CosineAnnealinLR &CosineAnnealinLR &CosineAnnealinLR &CosineAnnealinLR \\
Network hidden dimension & &(256, 256, 256, 256, 256, 256, 256) &(256, 256, 256, 256, 256, 256, 256) &(256, 256, 256, 256, 256, 256, 256) &(256, 256, 256, 256, 256, 256, 256) &(512, 256, 256, 512, 256, 256, 512) \\
Network blocks number & &(3,3,3,3,3,3,3) &(3,3,3,3,3,3,3) &(3,3,3,3,3,3,3) &(3,3,3,3,3,3,3) &(3,3,3,3,3,3,3) \\
Discount Factor & $\gamma$ & 0.99 & 0.99 & 0.99 & 0.99 & 0.99 \\

Data normalization & &\checkmark &\checkmark &\checkmark &\checkmark &\checkmark \\
Initial trajectory length & $l_{init}$ &(100, 100, 100, 200, 100, 100, 200) &(100, 100, 100, 200, 100, 100, 200) &(100, 100, 100, 200, 100, 100, 200) &(100, 100, 100, 200, 100, 100, 200) &(100, 100, 100, 200, 100, 100, 200) \\
Cliped trajectory length & $l_{\mathcal{D}^{(0)}}$ &(16,16,16,16,24,16,16) &(16,16,16,16,24,16,16) &(16,16,16,16,24,16,16) &(16,16,16,16,24,16,16) &(16,16,16,16,24,16,16) \\
Reference trajectory length & $l_{\mathcal{R}}$ & (500, 500, 500, 500, 500, 500, 500) & (500, 500, 500, 500, 500, 500, 500) & (500, 500, 500, 500, 500, 500, 500) & (500, 500, 500, 500, 500, 500, 500) & (500, 500, 500, 500, 500, 500, 500) \\

GMM cluster number & & - & 5 & - & - & - \\
\bottomrule
\end{tabular}
}
\label{tab: algo settings}
\end{center}
\end{sidewaystable}

The detailed dimensional information for the original and latent spaces for each task and algorithm is provided in \Cref{tab: algo settings}. For simplicity, we organize the hyperparameters corresponding to each task in an array format as follows: \texttt{(Flat-Anymal-D, Flat-Unitree-A1, Flat-Unitree-Go2, Rough-Unitree-Go2, Flat-Unitree-H1, Flat-Unitree-G1, Rough-Unitree-G1)}. We set a fixed step size $\Delta n$ for dimension increment at 100. The horizon length $H$ for loss computation and MPC solving is set to 24 for the Unitree-H1 robot and 16 for the others.

All networks implemented for testing the algorithms are Residual Neural Networks (\cite{he2015deepresiduallearningimage}), featuring residual blocks structured as \texttt{\{Linear\}-\{Relu\}-\{Linear\}-\{Residual\}-\{RELU\}}. The hidden dimensions and number of blocks are listed in \Cref{tab: algo settings}.  All networks are trained using the Adam optimizer with a Cosine Annealing learning rate scheduler. The initial learning rate and training epochs are 1e-3 and 100, respectively.

We normalize data in all test suites to mitigate the misaligned scale for each physical variable. The initial trajectories $m_{\mathcal{D}^{(0)}}$ is set to 6e6 with initial trajectories length $l_{\mathcal{D}^{(0)}}$ 100 for flat terrain and 200 for rough terrain. The clipped trajectory length $l_{init}$ is equal to horizon length $H$. Then we construct reference repository $\mathcal{R}$ with reference number $m_{\mathcal{R}}$ 3000 and reference length 500.

Other unique hyperparameters for each algorithm follow the original paper to attain the best performance.

Each model is trained on a server with a 48-core Intel(R) Xeon(R) Silver 6426Y CPU @ 2.5.GHz, four Nvidia RTX A6000 GPU with 48GB memory, and Ubuntu 22.04.

\subsection{Definition of Tracking Metrics}
\label{sec: def of metrics}
\begin{itemize}
    \item Joint-relative mean per-joint position error($E_{JrPE}$): $\frac{1}{200*J}\sum_{t=1}^{200}\|j_t - j_t^*\|_1$, where $j_t$ is the measured DoF position, $j_t^*$ is the reference DoF position, $J$ is the number of DoF.
    \item Joint-relative mean per-joint velocity error ($E_{JrVE}$): $\frac{1}{200*J}\sum_{t=1}^{200}\|\dot{j}_t - \dot{j}_t^*\|_1$
    \item Joint-relative mean per-joint acceleration error ($E_{JrAE}$): $\frac{1}{200*J}\sum_{t=1}^{200}\|\ddot{j}_t - \ddot{j}_t^*\|_1$
    \item Root mean position error ($E_{RPE}$): $\frac{1}{200*3}\sum_{t=1}^{200}\|p_t - p_t^*\|_1$, where $p_t$ is the measured root position, $p_t^*$ is the reference root position.
    \item Root mean linear velocity error ($E_{RLVE}$): $\frac{1}{200*3}\sum_{t=1}^{200}\|\dot{p}_t - \dot{p}_t^*\|_1$
    \item Root mean orientation error ($E_{ROE}$): $\frac{1}{200*4}\sum_{t=1}^{200}\|r_t - r_t^*\|_1$, where $r_t$ is the measured root orientation (represented as a quaternion), $r_t^*$ is the reference root orientation (represented as a quaternion).
    \item Root mean angular velocity error ($E_{RLAE}$): $\frac{1}{200*3}\sum_{t=1}^{200}\|\dot{r}_t - \dot{r}_t^*\|_1$
\end{itemize}

\subsection{Tracking Metrics for Each Task}
\label{sec: total tracking exp}

\begin{table*}[htbp]
\begin{center}
\begin{tabular}{c|ccc|cccc|c}
\toprule
\multicolumn{9}{c}{General Tracking Metrics}\\
\hline \\[-0.95em]
\multicolumn{1}{c|}{Algorithm}&\multicolumn{3}{c}{Joint-relative $\downarrow$} & \multicolumn{4}{|c|}{Root-relative $\downarrow$}& \multicolumn{1}{c}{Survival $\uparrow$}\\
\cline{2-9}\\[-1.02em]
 & $E_{JrPE}$ & $E_{JrVE}$ & $E_{JrAE}$ & $E_{RPE}$ & $E_{ROE}$ & $E_{RLVE}$ & $E_{RAVE}$ & $T_{Sur}$ \\
\hline \\[-0.95em]
\textbf{Ours} & \textbf{0.0467} & \textbf{0.9227} & 55.7386 & \textbf{0.1464} & \textbf{0.0418} & \textbf{0.1322} & \textbf{0.3367} & \textbf{194.5080} \\
DKRL & 0.0794 & 1.2533 & 69.3621 & 0.2752 & 0.0549 & 0.2077 & 0.4619 & 182.3020 \\
DKAC & 0.2422 & 1.7859 & 69.4619 & 0.5077 & 0.2532 & 0.3533 & 0.7922 & 44.6020 \\
DKUC & 0.3137 & 1.4798 & \textbf{47.1150} & 0.5221 & 0.3251 & 0.3472 & 0.7125 & 27.9680 \\
NNDM & 0.1841 & 1.8721 & 75.2555 & 0.5357 & 0.2126 & 0.3564 & 0.7197 & 35.0360 \\
\bottomrule
\end{tabular}
\caption{The average tracking metrics evaluated for each algorithm on Flat-Anymal-D.}
\label{tab: tracking performance for anymal-D}
\end{center}
\end{table*}

\begin{table*}[htbp]
\begin{center}
\begin{tabular}{c|ccc|cccc|c}
\toprule
\multicolumn{9}{c}{General Tracking Metrics}\\
\hline \\[-0.95em]
\multicolumn{1}{c|}{Algorithm}&\multicolumn{3}{c}{Joint-relative $\downarrow$} & \multicolumn{4}{|c|}{Root-relative $\downarrow$}& \multicolumn{1}{c}{Survival $\uparrow$}\\
\cline{2-9}\\[-1.02em]
 & $E_{JrPE}$ & $E_{JrVE}$ & $E_{JrAE}$ & $E_{RPE}$ & $E_{ROE}$ & $E_{RLVE}$ & $E_{RAVE}$ & $T_{Sur}$ \\
\hline \\[-0.95em]
\textbf{Ours} & \textbf{0.0322} & \textbf{0.8212} & \textbf{59.6711} & \textbf{0.0930} & \textbf{0.0256} & \textbf{0.0800} & \textbf{0.2657} & \textbf{195.4880} \\
DKRL & 0.0498 & 1.0385 & 71.4132 & 0.1180 & 0.0401 & 0.1039 & 0.3671 & 161.4340 \\
DKAC & 0.1462 & 3.8430 & 276.9420 & 0.3574 & 0.1101 & 0.2874 & 1.3771 & 10.8920 \\
DKUC & 0.0382 & 0.9525 & 69.6217 & 0.0997 & 0.0287 & 0.0896 & 0.3132 & 176.9560 \\
NNDM & 0.2166 & 3.9797 & 253.8551 & 0.4614 & 0.1215 & 0.3371 & 1.1712 & 3.8280 \\
\bottomrule
\end{tabular}
\caption{The average tracking metrics evaluated for each algorithm on Flat-Unitree-A1.}
\label{tab: tracking performance for a1}
\end{center}
\end{table*}

\begin{table*}[htbp]
\begin{center}
\begin{tabular}{c|ccc|cccc|c}
\toprule
\multicolumn{9}{c}{General Tracking Metrics}\\
\hline \\[-0.95em]
\multicolumn{1}{c|}{Algorithm}&\multicolumn{3}{c}{Joint-relative $\downarrow$} & \multicolumn{4}{|c|}{Root-relative $\downarrow$}& \multicolumn{1}{c}{Survival $\uparrow$}\\
\cline{2-9}\\[-1.02em]
 & $E_{JrPE}$ & $E_{JrVE}$ & $E_{JrAE}$ & $E_{RPE}$ & $E_{ROE}$ & $E_{RLVE}$ & $E_{RAVE}$ & $T_{Sur}$ \\
\hline \\[-0.95em]
\textbf{Ours} & \textbf{0.0428} & \textbf{0.9563} & \textbf{67.1742} & \textbf{0.1127} & \textbf{0.0364} & \textbf{0.0934} & \textbf{0.2946} & \textbf{195.1240} \\
DKRL & 0.0734 & 1.5231 & 104.0327 & 0.1817 & 0.0557 & 0.1432 & 0.4822 & 159.8200 \\
DKAC & 0.1710 & 2.9952 & 186.1956 & 0.3994 & 0.1956 & 0.2931 & 1.0813 & 19.5120 \\
DKUC & 0.1328 & 1.6218 & 89.0778 & 0.2163 & 0.1116 & 0.1814 & 0.5298 & 115.7020 \\
NNDM & 0.1716 & 3.3656 & 203.0976 & 0.4645 & 0.1290 & 0.3181 & 1.0053 & 4.6600 \\
\bottomrule
\end{tabular}
\caption{The average tracking metrics evaluated for each algorithm on Flat-Unitree-Go2.}
\label{tab: tracking performance for go2}
\end{center}
\end{table*}

\begin{table*}[htbp]
\begin{center}
\begin{tabular}{c|ccc|cccc|c}
\toprule
\multicolumn{9}{c}{General Tracking Metrics}\\
\hline \\[-0.95em]
\multicolumn{1}{c|}{Algorithm}&\multicolumn{3}{c}{Joint-relative $\downarrow$} & \multicolumn{4}{|c|}{Root-relative $\downarrow$}& \multicolumn{1}{c}{Survival $\uparrow$}\\
\cline{2-9}\\[-1.02em]
 & $E_{JrPE}$ & $E_{JrVE}$ & $E_{JrAE}$ & $E_{RPE}$ & $E_{ROE}$ & $E_{RLVE}$ & $E_{RAVE}$ & $T_{Sur}$ \\
\hline \\[-0.95em]
\textbf{Ours} & \textbf{0.0425} & \textbf{0.7963} & 55.5985 & \textbf{0.1001} & \textbf{0.1227} & \textbf{0.1021} & \textbf{0.4320} & \textbf{191.7280} \\
DKRL & 0.0846 & 1.5861 & 116.1124 & 0.1843 & 0.2604 & 0.1554 & 0.6729 & 71.1740 \\
DKAC & 0.1648 & 1.7733 & 91.0318 & 0.3195 & 0.4493 & 0.2015 & 0.7549 & 21.8420 \\
DKUC & 0.1177 & 0.8725 & \textbf{35.6163} & 0.1687 & 0.2227 & 0.1492 & 0.5043 & 82.0040 \\
NNDM & 0.1180 & 2.2806 & 147.1042 & 0.3023 & 0.4345 & 0.1997 & 0.8586 & 28.0060 \\
\bottomrule
\end{tabular}
\caption{The average tracking metrics evaluated for each algorithm on Rough-Unitree-Go2.}
\label{tab: tracking performance for go2 rough}
\end{center}
\end{table*}

\begin{table*}[htbp]
\begin{center}
\begin{tabular}{c|ccc|cccc|c}
\toprule
\multicolumn{9}{c}{General Tracking Metrics}\\
\hline \\[-0.95em]
\multicolumn{1}{c|}{Algorithm}&\multicolumn{3}{c}{Joint-relative $\downarrow$} & \multicolumn{4}{|c|}{Root-relative $\downarrow$}& \multicolumn{1}{c}{Survival $\uparrow$}\\
\cline{2-9}\\[-1.02em]
 & $E_{JrPE}$ & $E_{JrVE}$ & $E_{JrAE}$ & $E_{RPE}$ & $E_{ROE}$ & $E_{RLVE}$ & $E_{RAVE}$ & $T_{Sur}$ \\
\hline \\[-0.95em]
\textbf{Ours} & \textbf{0.0377} & \textbf{0.2983} & \textbf{15.1954} & \textbf{0.1860} & \textbf{0.0874} & \textbf{0.1949} & \textbf{0.3396} & \textbf{181.4740} \\
DKRL & 0.0992 & 0.5034 & 17.7740 & 0.3657 & 0.2438 & 0.2538 & 0.5304 & 98.7680 \\
DKAC & 0.1951 & 0.9522 & 32.7991 & 0.3813 & 0.3077 & 0.2613 & 0.7258 & 40.0660 \\
DKUC & 0.1925 & 0.9012 & 32.0362 & 0.3701 & 0.2438 & 0.2563 & 0.6709 & 71.2460 \\
NNDM & 0.1105 & 0.8381 & 31.3497 & 0.4377 & 0.2752 & 0.2792 & 0.5807 & 109.2700 \\
\bottomrule
\end{tabular}
\caption{The average tracking metrics evaluated for each algorithm on Flat-Unitree-H1.}
\label{tab: tracking performance for h1}
\end{center}
\end{table*}

\begin{table*}[htbp]
\begin{center}
\begin{tabular}{c|ccc|cccc|c}
\toprule
\multicolumn{9}{c}{General Tracking Metrics}\\
\hline \\[-0.95em]
\multicolumn{1}{c|}{Algorithm}&\multicolumn{3}{c}{Joint-relative $\downarrow$} & \multicolumn{4}{|c|}{Root-relative $\downarrow$}& \multicolumn{1}{c}{Survival $\uparrow$}\\
\cline{2-9}\\[-1.02em]
 & $E_{JrPE}$ & $E_{JrVE}$ & $E_{JrAE}$ & $E_{RPE}$ & $E_{ROE}$ & $E_{RLVE}$ & $E_{RAVE}$ & $T_{Sur}$ \\
\hline \\[-0.95em]
\textbf{Ours} & \textbf{0.0065} & \textbf{0.2346} & \textbf{17.1064} & \textbf{0.0219} & \textbf{0.0126} & \textbf{0.0365} & \textbf{0.1174} & \textbf{198.1140} \\
DKRL & 0.0858 & 1.1325 & 55.2168 & 0.5004 & 0.2182 & 0.3082 & 0.6091 & 73.2240 \\
DKAC & 0.1819 & 2.0094 & 110.0289 & 0.3974 & 0.3332 & 0.3274 & 0.9012 & 14.0400 \\
DKUC & 0.1579 & 0.9549 & 44.7592 & 0.3077 & 0.2420 & 0.2746 & 0.6048 & 51.1280 \\
NNDM & 0.1158 & 1.5910 & 87.3439 & 0.4060 & 0.2896 & 0.2994 & 0.7254 & 37.4180 \\
\bottomrule
\end{tabular}
\caption{The average tracking metrics evaluated for each algorithm on Flat-Unitree-G1.}
\label{tab: tracking performance for g1}
\end{center}
\end{table*}

\begin{table*}[htbp]
\begin{center}
\begin{tabular}{c|ccc|cccc|c}
\toprule
\multicolumn{9}{c}{General Tracking Metrics}\\
\hline \\[-0.95em]
\multicolumn{1}{c|}{Algorithm}&\multicolumn{3}{c}{Joint-relative $\downarrow$} & \multicolumn{4}{|c|}{Root-relative $\downarrow$}& \multicolumn{1}{c}{Survival $\uparrow$}\\
\cline{2-9}\\[-1.02em]
 & $E_{JrPE}$ & $E_{JrVE}$ & $E_{JrAE}$ & $E_{RPE}$ & $E_{ROE}$ & $E_{RLVE}$ & $E_{RAVE}$ & $T_{Sur}$ \\
\hline \\[-0.95em]
\textbf{Ours} & \textbf{0.0351} & \textbf{0.5202} & \textbf{31.5414} & \textbf{0.2017} & \textbf{0.1408} & \textbf{0.2120} & \textbf{0.5166} & \textbf{162.7240} \\
DKRL & 0.1039 & 0.8387 & 48.7527 & 0.4591 & 0.2198 & 0.2903 & 0.8203 & 71.9560 \\
DKAC & 0.1699 & 1.1268 & 56.4011 & 0.4061 & 0.2754 & 0.2979 & 0.7680 & 24.2240 \\
DKUC & 0.1503 & 0.7970 & 35.7952 & 0.3692 & 0.2186 & 0.2779 & 0.5559 & 52.2300 \\
NNDM & 0.0905 & 1.4870 & 94.1604 & 0.4262 & 0.2918 & 0.3075 & 0.9143 & 30.0780 \\
\bottomrule
\end{tabular}
\caption{The average tracking metrics evaluated for each algorithm on Rough-Unitree-G1.}
\label{tab: tracking performance for g1 rough}
\end{center}
\end{table*}

\end{document}